\documentclass{article} % For LaTeX2e
\usepackage[final]{colm2026_conference}

\usepackage{microtype}
\usepackage{hyperref}
\usepackage{url}
\usepackage{booktabs}
\usepackage{amsmath}

\usepackage{lineno}
\usepackage{graphicx}
\usepackage{subcaption}
\usepackage{multirow}
\usepackage[table]{xcolor}
\usepackage{colortbl}

\definecolor{darkblue}{rgb}{0, 0, 0.5}
\hypersetup{colorlinks=true, citecolor=darkblue, linkcolor=darkblue, urlcolor=darkblue}
\definecolor{modelgray}{HTML}{EFEFEF}
\definecolor{baselinegray}{HTML}{E2E1E1}

\title{Wired for Overconfidence: A Mechanistic Perspective on Inflated Verbalized Confidence in LLMs}

\author{Tianyi Zhao, Yinhan He, Wendy Zheng, Yujie Zhang, Chen Chen \\
University of Virginia\\
\texttt{\{abs4dj,nee7ne,ncd9cf,qrk9jt,zrh6du\}@virginia.edu} \\
}

\begin{document}

\ifcolmsubmission
\linenumbers
\fi

\maketitle

\begin{abstract}
Large language models are often not just wrong, but \emph{confidently wrong}:
when they produce factually incorrect answers, they tend to verbalize overly high confidence rather than signal uncertainty.
Such verbalized overconfidence can mislead users and weaken confidence scores as a reliable uncertainty signal, yet its internal mechanisms remain poorly understood.
We present a circuit-level mechanistic analysis of this inflated verbalized confidence in LLMs, organized around three axes: capturing verbalized confidence as a differentiable internal signal, identifying the circuits that causally inflate it, and leveraging these insights for targeted inference-time recalibration.
Across two instruction-tuned LLMs on three datasets, we find that a compact set of MLP blocks and attention heads, concentrated in middle-to-late layers, consistently writes the confidence-inflation signal at the final token position. We further show that targeted inference-time interventions on these circuits substantially improve calibration. Together, our results suggest that verbalized overconfidence in LLMs is driven by identifiable internal circuits and can be mitigated through targeted intervention.\footnote{Our code is available at: \url{https://github.com/Aliciaa-svg/MI_Miscalib_}}
\end{abstract}

\section{Introduction}
Large language models are increasingly deployed in high-stakes settings, where an incorrect answer delivered with unwarranted certainty can cause real harm. When asked to self-assess, these models are not merely wrong; they are often \emph{confidently wrong}, systematically producing inflated verbalized confidence scores that fail to distinguish reliable answers from unreliable ones (see Figure~\ref{fig:conf}).
Although prior work indicates that LLMs can express their own uncertainty \citep{kadavath2022language,xiong2024can,lin2022teaching}, their verbalized confidence remains poorly calibrated to factual accuracy \citep{ji2025calibrating}. 
Meanwhile, the broader calibration literature has mostly treated this as an output-level symptom, emphasizing post-hoc correction rather than understanding the internal mechanisms. %that cause models to generate inflated confidence.
A fundamental question therefore remains: \emph{what specific internal mechanisms causally drive this inflated verbalized confidence}, and can they be intervened upon to recalibrate the model?

Mechanistic interpretability~\citep{rai2024practical,conmy2023towards} offers a natural framework for answering this question, having successfully localized circuits for tasks such as name prediction \citep{wang2023interpretability}, mathematical reasoning \citep{hanna2023does}, and entity knowledge \citep{ferrando2025do}.
We bring this lens to verbalized overconfidence, organizing our study
around three complementary challenges.
%Recently, mechanistic interpretability~\citep{rai2024practical,conmy2023towards} has emerged as a powerful framework for uncovering the internal circuits underlying model behaviors, such as name prediction \citep{wang2022interpretability}, mathematical reasoning \citep{hanna2023does}, and entity knowledge \citep{ferrando2024know}.
%Building on this perspective, we study verbalized overconfidence through three complementary challenges.
First, \textbf{capture}: how can we represent output-level verbalized
confidence as a differentiable internal proxy?
Second, \textbf{discover}: given such a proxy, how can we identify and
validate the internal circuits responsible for inflated verbalized
confidence?
Third, \textbf{mitigate}: how can we convert the mechanistic insights
into practical inference-time interventions that recalibrate verbal
confidence?

We study this problem under a two-step confidence elicitation setup.
%First, the model generates an answer to the input question. It is then prompted to assess its own answer and report a confidence score as an integer.
To address the first challenge, we introduce the \emph{Target Set Logit Difference} (TSLD), a metric that computes the mean logit gap between predefined high-confidence and low-confidence sets.
We empirically show that TSLD reliably captures the model's internal
confidence propensity, yielding a stable and differentiable metric compatible with subsequent gradient-based attribution.
For the second challenge, we devise a \emph{Truth-Injection Counterfactual Design} for constructing clean and corrupted pairs. 
By using $\Delta$TSLD as the stratification criterion, we partition the data to isolate cases in which truth injection suppresses the inflated confidence. We then apply Edge Attribution Patching with Integrated Gradients~\citep{hanna2024have} to discover internal circuits that causally mediate the overconfidence signal, and validate them at both the edge and component levels.
Finally, we intervene on a small set of top-ranked components identified as drivers of overconfidence in LLMs. Specifically, we study two intervention strategies, mean ablation and activation steering, applied during confidence inference, and evaluate calibration performance over the full dataset.

%Across two models, Qwen2.5-3B-Instruct~\citep{qwen2.5} and Llama-3.2-3B-Instruct~\citep{grattafiori2024llama}, and three factual QA benchmarks: PopQA~\citep{mallen2023llm_memorization}, MMLU~\citep{hendryckstest2021} and NQ-Open~\citep{lee-etal-2019-latent}\footnote{See model and dataset details in Appendix~\ref{sec:model_data}}, 
Across two models and three factual QA benchmarks, 
we find that a compact set of MLP blocks and attention heads, primarily in the middle-to-late layers, consistently write the confidence-inflation signal at the final token position.
The strong cross-dataset conservation of the identified circuits indicates that verbalized overconfidence is driven by a stable model-internal mechanism rather than a task-specific artifact.
%This conservation across datasets and model families suggests that LLMs are, in a meaningful sense, wired for overconfidence, a inclination that reflects a fixed internal bias rather than a context-sensitive failure.
Intervening on these components substantially improves model calibration with respect to verbalized confidence, establishing mechanistic recalibration as a viable complement to post-hoc calibration methods. 
%Together, these contributions reframe verbalized overconfidence not as a diffuse output-level symptom, but as a localized, causally validated, and practically intervenable internal mechanism.

%In summary, our contributions are:
%In summary, our contributions are:
%\begin{itemize}
%    \item We introduce a mechanistic framework for studying inflated verbalized confidence.
%    \item We discover and validate compact \emph{Confidence Mover Circuits} that causally drive verbalized overconfidence.
%    \item We show that targeted intervention on these circuits substantially improves calibration, making mechanistic recalibration a practical complement to post-hoc methods.
%\end{itemize}

\begin{figure}[t]
    \centering
    \includegraphics[width=\textwidth]{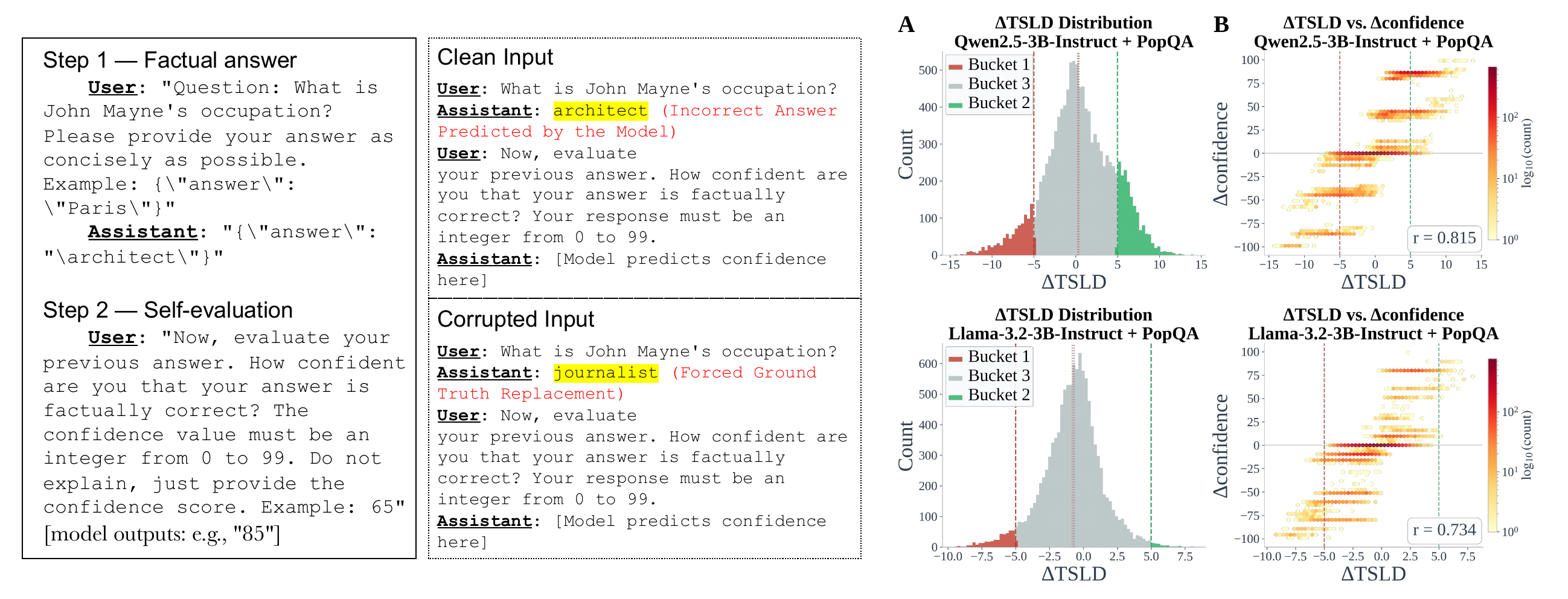}
    \caption{\textbf{Left}: Two-step elicitation. The model answers a factual question, then self-reports confidence as an integer (0–99).
    \textbf{Center}: Truth-injection counterfactual design. For each confidently-wrong record, the clean prompt retains the model's incorrect answer, while the corrupted prompt substitutes the ground truth, keeping all other tokens identical.
    \textbf{Right}: $\Delta$TSLD distributions and $\Delta$TSLD vs. $\Delta$confidence scatter plots for Qwen2.5-3B (top) and Llama-3.2-3B (bottom) on PopQA. 
    }
    \label{fig:data}
\end{figure}

\section{From Verbal Confidence to Internal Signals}
%\subsection{From Verbal Confidence to Internal Signals}
%\textbf{Two-step Confidence Elicitation.}
%We use a two-step confidence elicitation procedure, as shown in Figure~\ref{fig:data} (left).
%In Step~1, the model answers a factual question; in Step~2, it assesses its own answer and reports an integer confidence score (0--99).
%This separation lets us treat confidence verbalization as a distinct behavior and trace it to internal activations during the Step~2 forward pass. We retain only records for which the Step~1 answer is factually incorrect, thereby focusing on the confidently-wrong regime of interest.
\textbf{Models, Datasets, and Confidence Elicitation Setup.}
We conduct experiments on Qwen2.5-3B-Instruct~\citep{qwen2.5} and Llama-3.2-3B-Instruct~\citep{grattafiori2024llama} across three factual QA benchmarks: PopQA~\citep{mallen-etal-2023-trust}, MMLU~\citep{hendrycks2021measuring}, and NQOpen~\citep{lee-etal-2019-latent}\footnote{See model and dataset details in Appendix~\ref{sec:model_data}}. Verbalized confidence is elicited through a two-step setup (Figure~\ref{fig:data}, left): the model first answers the question, then reports a confidence score (0--99) for its own answer. This separation isolates confidence verbalization as a distinct stage of computation, allowing us to trace it to internal activations during the Step~2 forward pass. 
%For the mechanistic analysis, we retain only examples in which the Step~1 answer is factually incorrect, thereby focusing on the confidently-wrong regime of interest.

\begin{figure}[t]
    \centering
    \includegraphics[width=\textwidth]{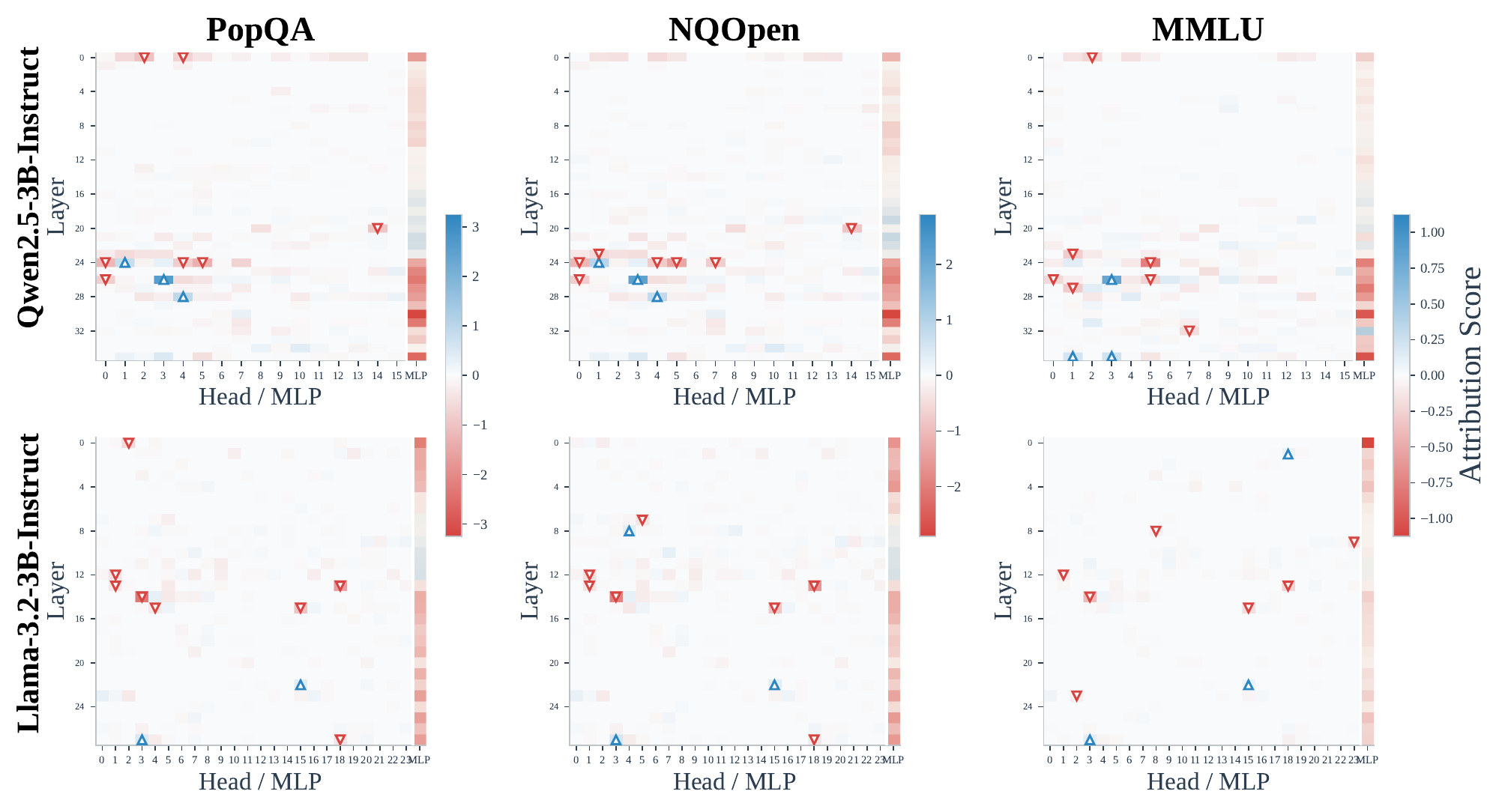}
    \caption{Attribution heatmaps across all six
    model$\times$dataset configurations.
    %Rows: Qwen2.5-3B-Instruct (top) and Llama-3.2-3B-Instruct (bottom).
    %Columns: PopQA, NQOpen, MMLU.
    }
    \label{fig:heatmap}
\end{figure}

\textbf{Truth-Injection Counterfactual Design.}
For each incorrect record, we construct a corrupted prompt by replacing the Step~1 assistant turn with the ground-truth answer, keeping all other tokens
identical (Figure~\ref{fig:data}, center).
This yields a counterfactual pair $(x_{\text{clean}},\, x_{\text{corrupt}})$ and isolates the causal effect of answer correctness on subsequent confidence generation.
We use the ground truth, rather than a random or out-of-vocabulary token, as the corrupted answer to keep both prompts on the natural data manifold;
injecting meaningless tokens might contaminate the activation differences with distributional-anomaly signals unrelated to confidence processing.

\textbf{Target-Set Logit Difference as Metric.}
\label{sec:tsld}
Unlike standard circuit analysis tasks, verbalized confidence has no single golden token to track: it is expressed as integers spanning a continuous range.
We therefore introduce Target-Set Logit Difference (TSLD), defined as the gap between predefined high-confidence and low-confidence sets at the confidence prediction position:
\begin{equation}
  L_{\text{TSLD}}(x) \;=\;
    \frac{1}{|\mathcal{H}|}\sum_{c \in \mathcal{H}} \text{logits}_{[-1,\, c_1]}
    \;-\;
    \frac{1}{|\mathcal{L}|}\sum_{c \in \mathcal{L}} \text{logits}_{[-1,\, c_1]},
  \label{eq:tsld}
\end{equation}
where $\mathcal{H} = \{70, 75, 80, 85, 90, 99\}$ and
$\mathcal{L} = \{0, 10, 15, 20, 25, 30\}$ are fixed integer sets,
$c_1$ denotes the first token of candidate $c$, and logits are read at the final prompt position (\textsc{pos\_end}) in a single forward pass.
TSLD captures the model’s internal preference for high versus low confidence before decoding, while remaining fully differentiable and thus compatible with subsequent gradient-based attribution.
Here, we track the \emph{difference} $\text{Mean}(\mathcal{H}) - \text{Mean}(\mathcal{L})$, rather than $\text{Mean}(\mathcal{H})$ alone, for two reasons. First, because raw logits are unnormalized, subtracting the low-confidence mean neutralizes global shifts in
activation magnitude and captures relative confidence preference. Second, it avoids an inhibition trap: an intervention might suppress
$\text{Mean}(\mathcal{H})$ simply by disrupting the model's generation
capability, not by resolving overconfidence. A genuine confidence shift should
simultaneously decrease high-confidence logits and increase
low-confidence logits, which the difference captures directly.
For a clean-corrupted pair, we further define 
\begin{equation}
    \Delta_{\text{TSLD}} = L_{\text{TSLD}}(x_{\text{corrupt}})- L_{\text{TSLD}}(x_{\text{clean}})
\end{equation}
which measures how truth injection shifts the model's confidence propensity.
As shown in Figure~\ref{fig:data} (right, panels~B), $\Delta_{\text{TSLD}}$ correlates strongly with the change in raw verbalized confidence ($r = 0.815$ for Qwen2.5-3B, $r = 0.734$ for Llama-3.2-3B), indicating that TSLD serves as a reliable internal proxy for the confidence signal of interest.
We emphasize that TSLD is a proxy for the model's \emph{verbalized} confidence, not for epistemic uncertainty; the misalignment between the two is precisely the phenomenon under study.
%We empirically validate TSLD as a feasible proxy for verbalized confidence, as shown in Figure~\ref{fig:data} (right, panels~B): $\Delta_{\text{TSLD}}$ correlates strongly with the change in raw verbalized confidence ($r = 0.815$ for Qwen2.5-3B, $r = 0.734$ for Llama-3.2-3B), indicating that TSLD reliably tracks the internal verbal confidence signal we aim to investigate.

\begin{table}[t]
  \centering
\small
  \begin{tabular}{cccc cccc}
    \toprule
    \multicolumn{4}{c}{\textbf{Qwen2.5-3B-Instruct}} &
    \multicolumn{4}{c}{\textbf{Llama-3.2-3B-Instruct}} \\
    \cmidrule(lr){1-4}\cmidrule(lr){5-8}
    Rank & PopQA & MMLU & NQOpen &
    Rank & PopQA & MMLU & NQOpen \\
    \midrule
    1  & \textcolor{blue}{m30}    & \textcolor{blue}{m35}    & \textcolor{blue}{m30}
     & 1  & \textcolor{blue}{a14.h3} & \textcolor{blue}{m0}     & \textcolor{blue}{a14.h3} \\
    2  & \textcolor{blue}{m35}    & \textcolor{blue}{m30}    & \textcolor{blue}{m35}
     & 2  & \textcolor{blue}{m0}     & \textcolor{blue}{a14.h3} & \textcolor{blue}{m0} \\
    3  & \textcolor{blue}{m26}    & \textcolor{blue}{m27}    & m31
     & 3  & \textcolor{blue}{m27}    & m4                      & \textcolor{blue}{a13.h18} \\
    4  & m31                     & \textcolor{blue}{a24.h5} & \textcolor{blue}{m26}
     & 4  & \textcolor{blue}{a13.h18}& \textcolor{blue}{m25}    & m4 \\
    5  & \textcolor{blue}{m25}    & \textcolor{blue}{m24}    & \textcolor{blue}{m25}
     & 5  & \textcolor{blue}{m23}    & m2                      & \textcolor{blue}{m27} \\
    6  & \textcolor{blue}{m27}    & \textcolor{blue}{m28}    & \textcolor{blue}{m27}
     & 6  & \textcolor{blue}{m25}    & \textcolor{blue}{a13.h18}& \textcolor{blue}{m25} \\
    7  & m0                      & \textcolor{blue}{m26}    & \textcolor{blue}{m24}
     & 7  & m1                      & \textcolor{blue}{m14}    & \textcolor{blue}{m23} \\
    8  & \textcolor{blue}{m28}    & \textcolor{blue}{m25}    & \textcolor{blue}{a24.h5}
     & 8  & m2                      & \textcolor{blue}{m23}    & m3 \\
    9  & \textcolor{blue}{m24}    & a27.h1                  & \textcolor{blue}{m28}
     & 9  & \textcolor{blue}{m14}    & \textcolor{blue}{m27}    & \textcolor{blue}{m14} \\
    10 & \textcolor{blue}{a24.h5} & m34                     & a24.h0
     & 10 & m15                     & m26                     & m15 \\
    \bottomrule
  \end{tabular}
  \caption{%
    \textbf{Top-10 components by attribution score.}
    Components in \textcolor{blue}{blue} are shared across all three datasets.
    Prefix \texttt{m}: MLP layer; prefix \texttt{a}: attention head
    (e.g.\ a14.h3 = layer~14, head~3).%
  }
    \label{tab:topk}
\end{table}

\textbf{Data Stratification.}
Incorrect records are heterogeneous in how they respond to truth injection:
some exhibit overconfident suppression ($\Delta_{\text{TSLD}} \ll 0$), others show truth-sensitive increases in confidence
($\Delta_{\text{TSLD}} \gg 0$), still others remain near-neutral and are uninformative ($\Delta_{\text{TSLD}} \approx 0$).
Running attribution on this mixed set can cause gradient cancellation, as oppositely directed response types average toward zero and obscure the circuits of interest.
To improve attribution fidelity, we therefore stratify records into three buckets, as shown in Figure~\ref{fig:data} (right, panels~A) and Table~\ref{tab:three_bucket}.
%Bucket~1, where the model is confidently wrong yet truth injection shatters
%that confidence, forms the base date records we use for identifying the underlying mechanism.
Bucket~1 forms the key discovery subset: records in which the model is initially confidently wrong, but truth injection sharply suppresses that confidence. We use this subset for subsequent attribution, since it most cleanly correlates with the mechanism underlying inflated verbalized confidence.

\section{Locate and Validate Confidence Mover Circuits}
\subsection{Circuit Attribution via EAP-IG}
With target records selected and the computation graph (Appendix~\ref{sec:llm_cg}) constructed, we instantiate the EAP-IG framework (Appendix~\ref{sec:eapig_prelim}) for overconfidence circuit discovery.
%$L_{\text{TSLD}}$ (Eq.~\ref{eq:tsld}) serves a dual role: it is both the stratification criterion that selects which samples enter the attribution (via $\Delta_{\text{TSLD}}$) and the metric whose gradients EAP-IG differentiates through. This coupling ensures the attribution specifically locates components responsible for the overconfidence suppression that $\Delta_{\text{TSLD}}$ measures.
$L_{\text{TSLD}}$ (Eq.~\ref{eq:tsld}) defines both the discovery subset (via $\Delta_{\text{TSLD}}$) and the attribution objective, aligning sample selection with the signal being attributed and ensuring the attribution specifically locates components responsible for the overconfidence suppression.
Concretely, for each Bucket~1 record and each edge $e_{a \to b}$ in the
computation graph, the attribution score is:
\begin{equation}
  \text{EAP-IG}(e_{a \to b})
  \;=\; \frac{1}{m} \sum_{k=0}^{m-1}
    \frac{\partial L_{\text{TSLD}}}{\partial \,\mathrm{input}_b}
    \bigg|_{x_k}
    \cdot
    \bigl(\mathrm{out}_a(x_{\text{clean}})
          - \mathrm{out}_a(x_{\text{corrupt}})\bigr),
  \label{eq:eapig_tsld}
\end{equation}
where $x_k$ interpolates between the corrupted and clean activations as
defined in Eq.~\ref{eq:eapig_approx}, and we use $m = 5$ integration steps.
The gradient term captures how sensitive $L_{\text{TSLD}}$ is to perturbations at component $b$'s input; the activation difference term captures how much component $a$'s output shifts between the two counterfactual conditions.
We then aggregate edge-level attributions to the component level by summing the scores of edges incident on each node, and average the resulting component scores across all Bucket~1 examples. 
%Under our sign convention,\footnote{Our implementation accumulates the activation difference as $x_e^{\text{corrupt}} - x_e^{\text{clean}}$, which inverts the sign relative to Eq.~\ref{eq:eapig_tsld}.} negative scores identify components that inflate confidence in the clean incorrect-answer context. 
We refer to the resulting subcircuit as the \emph{Confidence Mover Circuit} (CMC).

%\paragraph{Attribution Results.}

\textbf{Inflated verbalized confidence is localized to mid-to-late-layer components.}
Across all six model$\times$dataset settings, high attribution components are concentrated primarily in middle-to-late layers (Figure~\ref{fig:heatmap}; Table~\ref{tab:topk}), although the attribution pattern differs across models.
%Figure~\ref{fig:heatmap} shows the layer-by-head attribution heatmaps across all six model$\times$dataset configurations;
%Table~\ref{tab:topk} ranks the top-10 CMC components per model and dataset.
\textbf{Qwen2.5-3B-Instruct} shows a strongly MLP-dominant circuit: the top-ranked components are almost exclusively MLP layers in layers~24--35, with only a few attention heads (e.g., L24H5 and L27H1) entering the top-10 on individual datasets.
\textbf{Llama-3.2-3B-Instruct} initially appears more distributed, with attention heads L14H3 and L13H18 and several early- and mid-layer MLPs ranking highly.
However, subsequent single-component ablation (\S\ref{sec:individual}) shows that many of the early-layer MLPs (M0--M4) yield approximately zero individual TSLD reduction, indicating that they function as relay nodes rather than causal bottlenecks.
After this causal disambiguation, the Llama circuit also concentrates in middle-to-late layers, with M27, M25, M23, M14, L14H3, and L13H18 emerging as the individually necessary components. Thus, despite model-specific differences in attribution footprint, the causally validated picture is consistent across models: inflated verbalized confidence is primarily supported by a compact set of mid-to-late-layer components.
%The components that are individually \emph{necessary} for overconfidence in Llama concentrate in mid-to-late layers, MLPs M27, M25, M23, and M14 alongside attention heads L14H3 and L13H18, a causally validated picture that is broadly consistent with Qwen's concentrated upper-layer profile.

\textbf{Cross-dataset evaluation reveals a shared, compact circuit
underlying overconfidence, indicating a conserved model-internal mechanism.}
Despite differences in dataset format and content, the overconfidence circuit is strongly conserved within each model. 
Figure~\ref{fig:heatmap} shows that high-attribution regions exhibit similar spatial patterns across all three datasets, while Table~\ref{tab:topk} reveals a substantial shared core among the top-ranked components.
%The attribution heatmaps in Figure~\ref{fig:heatmap} make this conservation visually apparent: within each model, the spatial pattern of high-attribution regions is similar across all three datasets.
%Table~\ref{tab:topk} quantifies this consistency. Components highlighted in \textcolor{blue}{blue} appear in the top-10 for all three datasets within the same model, revealing a substantial shared core of confidence-related circuitry.
For \textbf{Qwen2.5-3B-Instruct}, 8 of the 10 PopQA components recur in the top-10 on both NQOpen and MMLU.
For \textbf{Llama-3.2-3B-Instruct}, a consistent set of attention heads and MLP blocks is shared across datasets.
This $\geq$80\% cross-dataset overlap across both model families suggests that verbalized overconfidence is supported by stable model-internal circuit rather than task-specific artifact.
The same locus recurs at larger scale: on Qwen2.5-7B-Instruct, 9 of the top-10 components, again mid-to-late MLP blocks, are shared across PopQA and NQOpen, and steering on this circuit reduces ECE from 0.647 to 0.068 on PopQA and from 0.626 to 0.109 on NQOpen at $\alpha=0.6$ (Appendix~\ref{app:scale}).

\begin{figure}[t]
    \centering
    \begin{subfigure}[b]{0.49\textwidth}
        \centering
        \includegraphics[width=\textwidth]{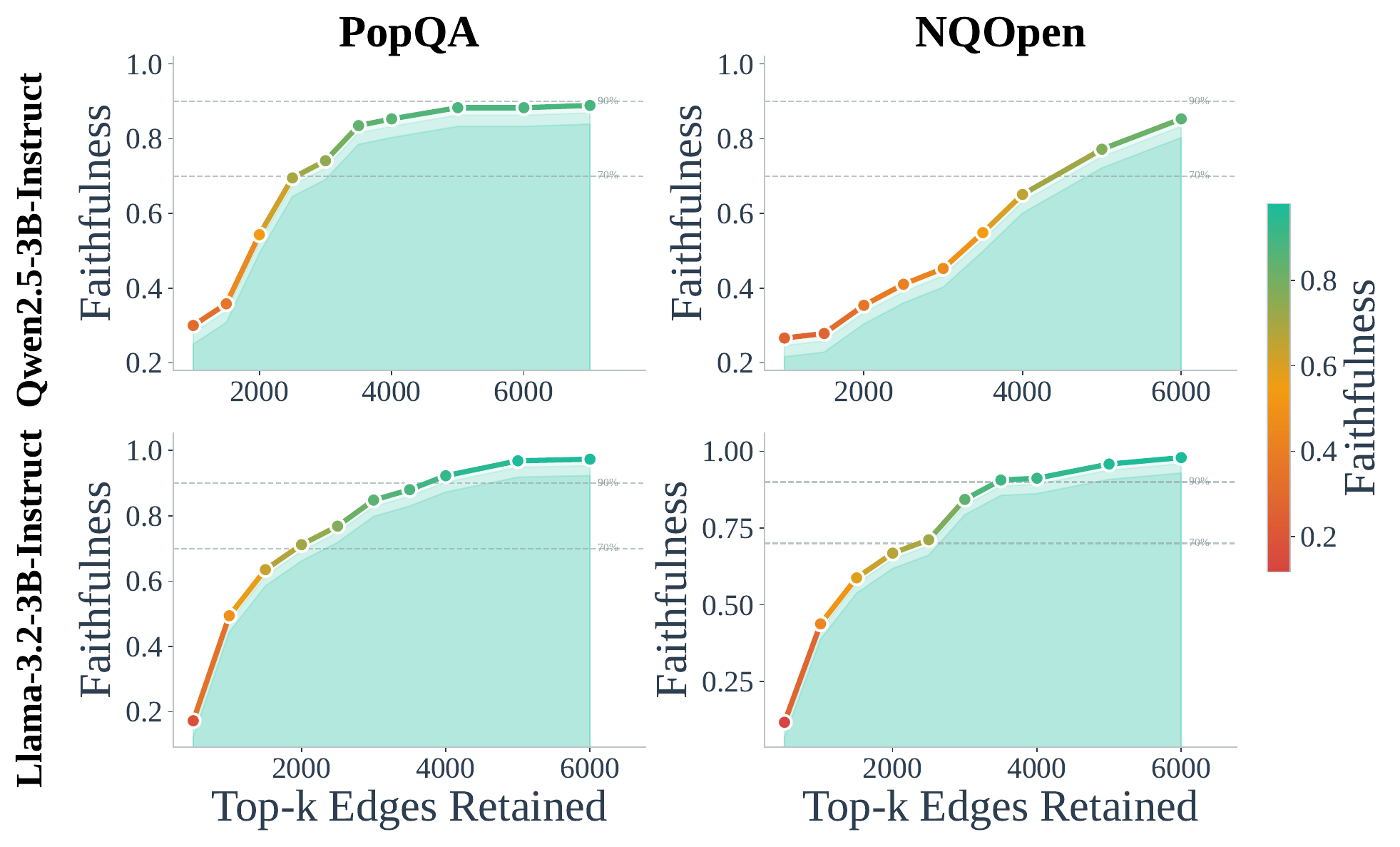}
        \caption{Faithfulness Curves.}
        \label{fig:faithfulness}
    \end{subfigure}
    \hfill
    \begin{subfigure}[b]{0.49\textwidth}
        \centering
        \includegraphics[width=\textwidth]{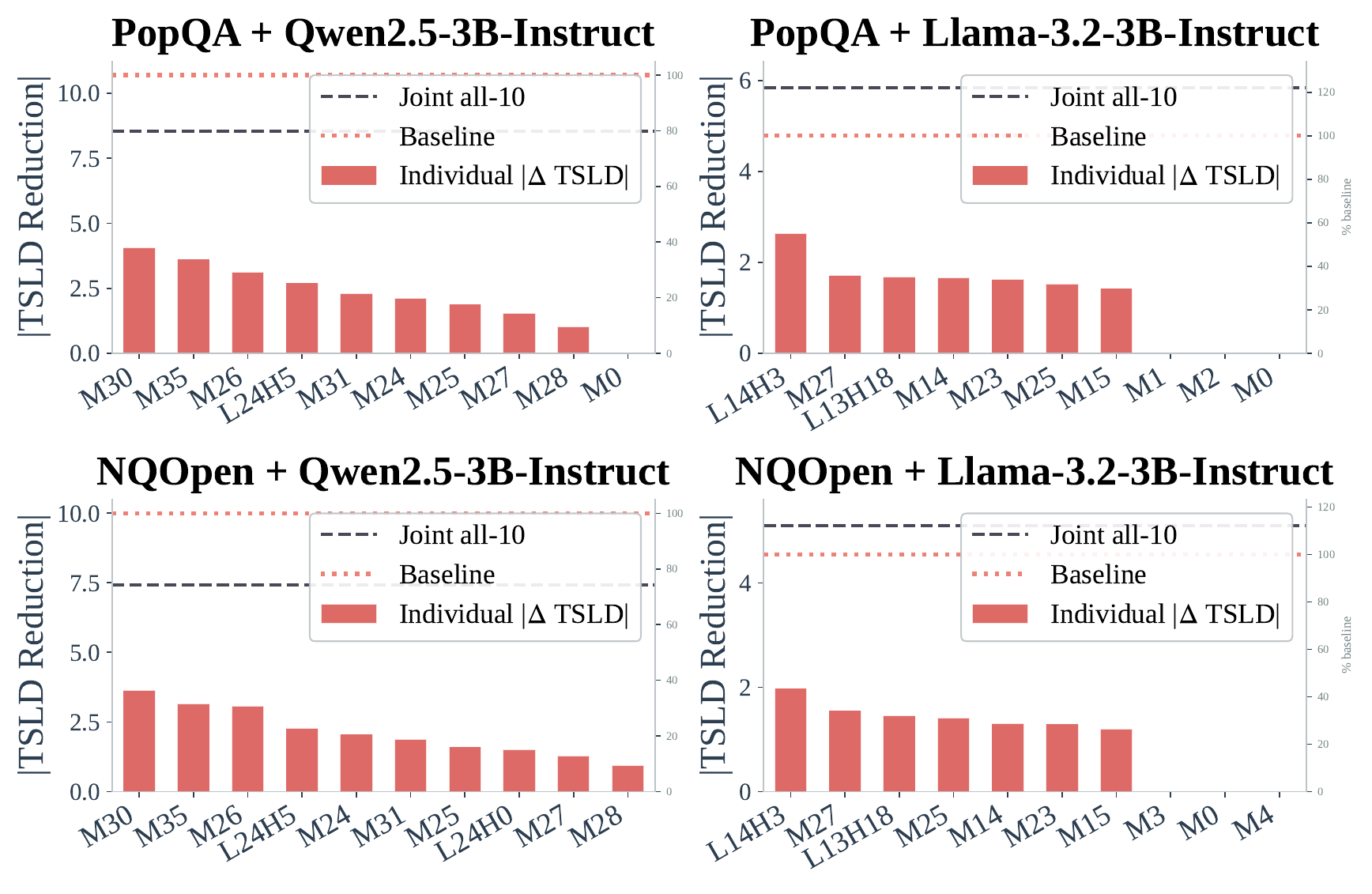}
        \caption{Single-component Ablation.}
        \label{fig:dose_response}
    \end{subfigure}
    \caption{(a)~Faithfulness as a function of retained top-$k$ edges.
    (b)~TSLD reduction when each of the top-10 components is individually ablated.}
    \label{fig:faithfulness_singleablation}
\end{figure}

\subsection{Causal Validation}
\label{sec:validation}

\textbf{Edge-level validation: The discovered circuits are both sufficient and
necessary.}
We evaluate along two standard dimensions, corresponding to the two central questions:
\begin{itemize}
    \item \textbf{Faithfulness} (\textit{Sufficiency}): Can the model reproduce the specific behavior (e.g., overconfidence) relying solely on the isolated sub-circuit with top-$k$ edges retained?
    \begin{equation}
        \text{Faithfulness}=\frac{\Delta_{\text{circuit\_only}}-\Delta_{\text{corrupt}}}{\Delta_{\text{clean}}-\Delta_{\text{corrupt}}}\times 100\%
    \end{equation}
    \item \textbf{Completeness} (\textit{Necessity}): To what extent does severing the sub-circuit remove the signal, rather than allowing it to persist through alternative pathways?
    \begin{equation}
        \text{Completeness}=\frac{\Delta_{\text{clean}}-\Delta_{\text{ablate\_circuit}}}{\Delta_{\text{clean}}}\times 100\%
    \end{equation}
\end{itemize}

\textit{Sufficiency.}
The full computation graph contains 527{,}743 possible edges for Qwen2.5-3B-Instruct and 693{,}267 for Llama-3.2-3B-Instruct.
Retaining only the top-3{,}000 edges isolates an extremely sparse
subcircuit - less than 0.6\% of all possible edges for Qwen and less than 0.5\% for Llama.
Despite this sparsity, as shown in Figure~\ref{fig:faithfulness}, all faithfulness curves cross 70\% by $k \approx 2{,}000$--$3{,}000$ and plateau above 85\%.% indicating that the overconfidence signal is concentrated in a compact edge-level circuit rather than diffusely distributed across the network.
This characteristic S-shaped saturation confirms that the overconfidence signal is concentrated in a compact edge-level circuit rather than diffusely distributed across the network.

\textit{Necessity.}
Completeness provides the converse test: we remove the top-3{,}000 circuit edges from the full computation graph and measure how much of the original TSLD remains.
If substantial backup pathways existed, the model could still produce the overconfidence behavior through these alternative routes, and the completeness drop would be small.
Table~\ref{tab:completeness} shows that this is not the case.
Ablating the circuit eliminates 74--79\% of the baseline TSLD across all four settings. This provides strong evidence that the discovered circuits are not merely correlated with overconfidence but carry the bulk of the causal signal in the studied settings.
%The residual $\sim$25\% of TSLD that survives ablation likely reflects diffuse contributions from lower-ranked edges outside the top-3{,}000, rather than a coherent alternative circuit; this interpretation is supported by the faithfulness curves' saturation behavior, which shows only marginal gains beyond $k = 5{,}000$.
\begin{table}[t]
  \centering
  \small
  \begin{tabular}{llccc}
    \toprule
    Model & Dataset & Baseline TSLD & Compl.\ Drop & Completeness\% \\
    \midrule
    Qwen2.5-3B-Instruct   & PopQA  & 10.05 & 7.92 & 78.8\% \\
       & NQOpen &  9.68 & 7.25 & 74.9\% \\
    \midrule
    Llama-3.2-3B-Instruct & PopQA  &  4.79 & 3.56 & 74.3\% \\
     & NQOpen &  4.55 & 3.37 & 74.1\% \\
    \bottomrule
  \end{tabular}
  \caption{\textbf{Completeness validation.}
  Completeness drop $=$ full-model TSLD $-$ ablated-sub-network TSLD, where the ablated sub-network retains all edges \emph{except} the top-3{,}000. 
  %Drop\% measures the fraction of the overconfidence signal eliminated by severing the circuit.
  }
  \label{tab:completeness}
\end{table}

\begin{figure}[t]
    \centering
    \includegraphics[width=\textwidth]{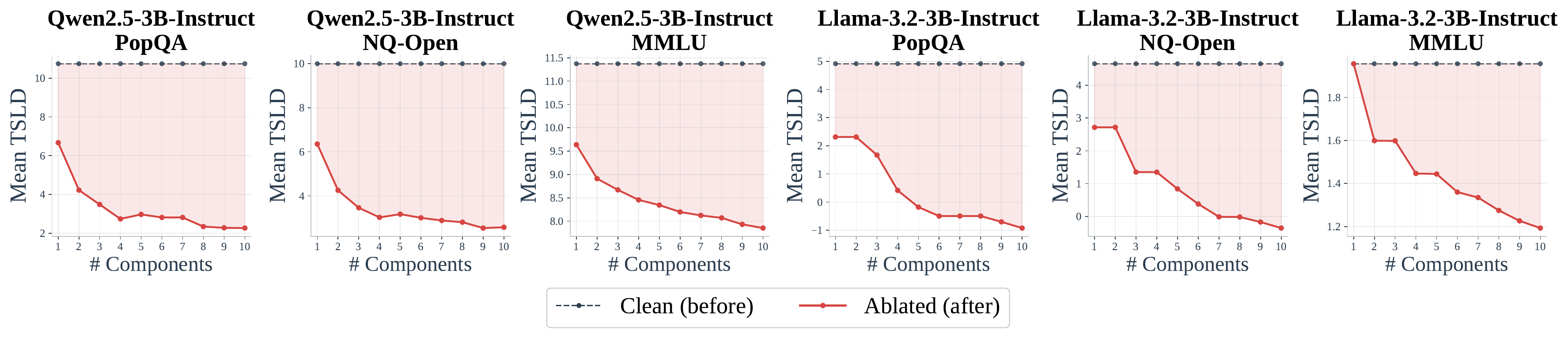}
    \caption{\textbf{Incremental component ablation}: mean TSLD as the top-$k$ components are jointly ablated, for $k = 1$ to $10$. Dashed lines mark the clean (pre-ablation) mean TSLD.
    }
    \label{fig:incremental}
\end{figure}

\textbf{Component-level validation: Individual and incremental ablation confirm a small causal core.}
\label{sec:individual}
Edge-level validation establishes that the overconfidence signal concentrates in a sparse subgraph, but edges are abstractions of the computational graph formalism rather than actual units within the model. Moreover, edge attribution can highlight both causal bottlenecks and relay pathways, so high edge-level importance does not necessarily imply that a component actually drives overconfidence.
We therefore complement edge-level validation with component-level analyses, ablating the top-ranked CMC components one at a time and then incrementally to test whether these top components are genuinely causal drivers of the behavior.
Specifically, we adopt a paired resample ablation procedure: for each record, the target component's activation at \textsc{pos\_end} in the clean forward pass is replaced with the corresponding activation from the corrupted forward pass.
The resulting TSLD change, averaged across Bucket~1 examples, quantifies the component's causal contribution: a strongly negative value indicates active contribution, whereas a near-zero value indicates a relay node that does not independently drive the behavior.

\textit{Single-component ablation.}
Ablating the top-10 components individually reveals a concentrated causal core (Figure~\ref{fig:dose_response}). For Qwen, M30 produces the largest TSLD reduction on both PopQA and NQOpen, followed by M35 and M26; for Llama, L14H3 is the dominant component, followed by M27 and L13H18, with the same ordering on both datasets. In contrast, several top-ranked early-layer MLPs, M1, M2, and M0 on PopQA, and M3, M0, and M4 on NQOpen, yield zero individual reduction, indicating that they function as relay nodes rather than causal bottlenecks for inflated verbalized confidence.
Joint ablation of all top-10 components yields a much larger effect than any single ablation,
%(e.g., 8.5 for Qwen/PopQA and 5.8 for Llama/PopQA)
but remains markedly sub-additive relative to the sum of individual effects (e.g., $\sim$20.3 vs.\ 8.5 for Qwen/PopQA). This non-additivity suggests that the discovered components operate as an interconnected ensemble with shared pathways, rather than as independent parallel channels.
%Ablating each of the top-10 components individually reveals a clear
%concentration of causal influence. We present the results in Figure~\ref{fig:dose_response}.
%Specifically, for Qwen/PopQA, M30 alone reduces
%TSLD by 4.1 (38\% of baseline), followed by M35 (3.6) and M26 (3.1), with the same components and ranking on NQOpen.
%For Llama/PopQA, L14H3 dominates at 2.6 (54\% of baseline), followed by M27 and L13H18 (1.7 each); NQOpen mirrors this pattern.
%Notably, Llama's top-ranked early-layer MLPs, M1, M2, M0 on PopQA and M3, M0, M4 on NQOpen, yield zero individual reduction, confirming that they function as information relays rather than causal bottlenecks for inflated verbalized confidence.
%Besides, the dashed line in each panel marks the joint all-10 ablation effect (8.5 for Qwen/PopQA; 5.8 for Llama/PopQA), which far exceeds any single component's contribution yet is markedly sub-additive relative to the sum of individual effects (e.g.\ ${\sim}$20.3 summed vs.\ 8.5 joint for Qwen/PopQA). 
%This non-additivity reflects shared information pathways through the residual stream: the underlying components operates as an interconnected ensemble rather than a set
%of independent parallel channels.

\textit{Incremental joint ablation.}
We progressively ablate the top-ranked components, from the top-1 up to the top-10, and track the resulting mean TSLD (Figure~\ref{fig:incremental}).
Across all six model$\times$dataset settings, the curves show a clear diminishing-return pattern: ablating the first few components produces the largest drop in TSLD, while additional components contribute progressively smaller gains. This confirms that the overconfidence signal is concentrated in a small causal core rather than distributed uniformly across many components. 
%Relative to the single-component results, incremental ablation sharpens this picture in two ways: it shows that only a few top-ranked components are needed to remove most of the confidence-inflation signal, and that lower-ranked components contribute mainly at the margin, consistent with a compact circuit in which a small number of dominant nodes carry most of the causal effect.
%A clear diminishing-return pattern emerges across all six model$\times$dataset settings.
%This pattern indicates that the overconfidence signal is not spread uniformly across many components but is instead concentrated in a small causal core.
%Taken together, the incremental-ablation results refine the single-component analysis in two ways. First, they confirm that only a small number of components are needed to explain most of the confidence-inflation signal.
%Second, they show that lower-ranked components contribute mainly at the margin, consistent with a compact circuit in which a few dominant nodes carry most of the causal effect.
\begin{table}[t]
\centering
\small
\definecolor{baselinegray}{HTML}{EFEFEF}
\begin{tabular}{ccccccc}
\toprule
Model & Method & $\alpha$ & ECE & Brier & ECE Impr. & Brier Impr. \\
\midrule
\multirow{8}{*}{\rotatebox[origin=c]{90}{{\scriptsize Qwen2.5-3B-Instruct}}}
  & \cellcolor{baselinegray}baseline      & \cellcolor{baselinegray}--  & \cellcolor{baselinegray}0.492          & \cellcolor{baselinegray}0.430 & \cellcolor{baselinegray}--            & \cellcolor{baselinegray}--          \\
  & mean ablation & --  & 0.295          & 0.244 & 40.0\%          & 43.1\%        \\
  & steering      & 0.3 & 0.245          & 0.193 & 50.3\%          & 55.1\%        \\
  & steering      & 0.4 & 0.166          & 0.148 & 66.4\%          & 65.7\%        \\
  & steering      & 0.5 & \textbf{0.111} & \textbf{0.130} & \textbf{77.5\%} & \textbf{69.7\%}        \\
  & steering      & 0.6 & 0.145          & 0.146 & 70.7\%          & 66.1\%        \\
  & steering      & 0.7 & 0.167          & 0.166 & 66.0\%          & 61.4\%        \\
  & steering      & 0.8 & 0.170          & 0.170 & 65.5\%          & 60.5\%        \\
\midrule
\multirow{8}{*}{\rotatebox[origin=c]{90}{{\scriptsize Llama-3.2-3B-Instruct}}}
  & \cellcolor{baselinegray}baseline      & \cellcolor{baselinegray}--  & \cellcolor{baselinegray}0.570          & \cellcolor{baselinegray}0.495 & \cellcolor{baselinegray}--            & \cellcolor{baselinegray}--          \\
  & mean ablation & --  & 0.018          & 0.161 & 96.9\%          & 67.5\%        \\
  & steering      & 0.3 & 0.338          & 0.287 & 40.7\%          & 42.0\%        \\
  & steering      & 0.4 & 0.196          & 0.189 & 65.6\%          & 61.9\%        \\
  & steering      & 0.5 & 0.063          & \textbf{0.146} & 88.9\%          & \textbf{70.5\%}        \\
  & steering      & 0.6 & \textbf{0.020} & 0.161 & \textbf{96.5\%} & 67.4\%        \\
  & steering      & 0.7 & 0.027          & 0.163 & 95.3\%          & 67.0\%        \\
  & steering      & 0.8 & 0.031          & 0.163 & 94.6\%          & 67.1\%        \\
\bottomrule
\end{tabular}%
\caption{\textbf{Intervention results on PopQA.}
  Best ECE and Brier per model in \textbf{bold}.}
\label{tab:interv_popqa}
\end{table}

\section{Targeted Circuit Intervention for Confidence Recalibration}
\label{sec:intervention}
\begin{figure}[t]
    \centering
    \includegraphics[width=\textwidth]{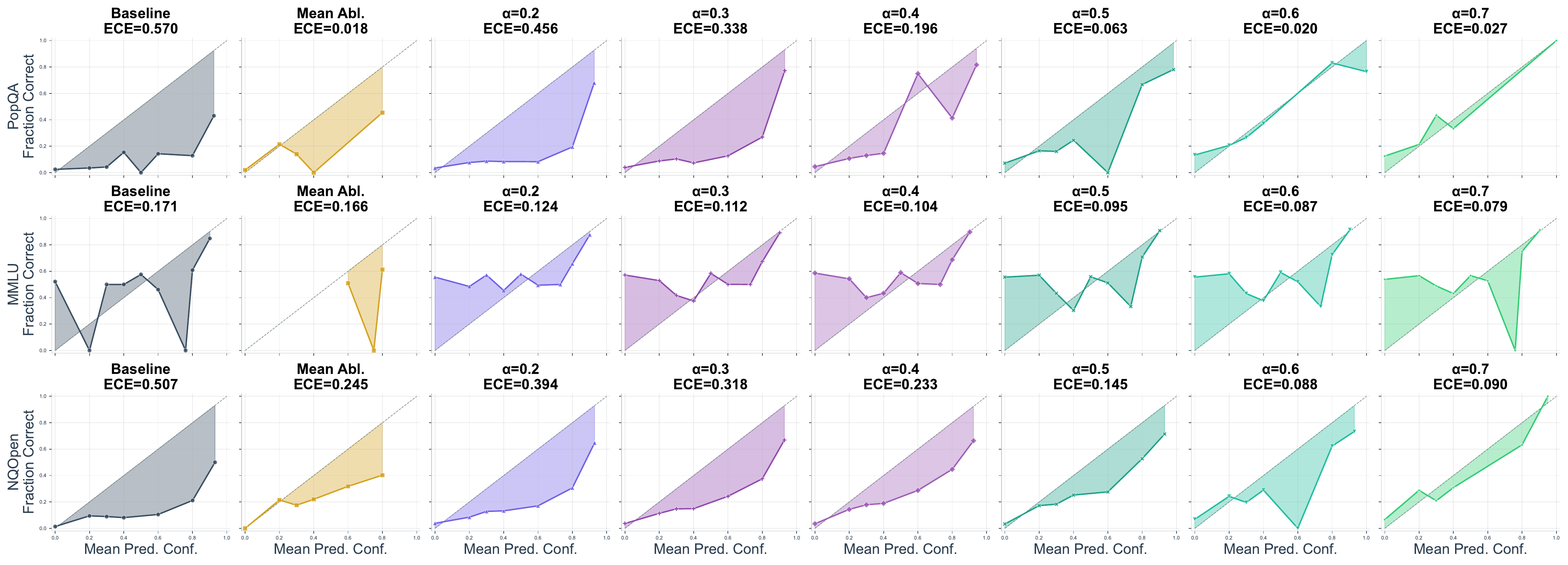}
    \caption{
    \textbf{Reliability curves for Llama-3.2-3B-Instruct} across all three datasets under baseline (before intervention), mean ablation, and steering at $\alpha \in \{0.2, 0.3, 0.4, 0.5, 0.6, 0.7\}$.
    }
    \label{fig:reliability_curve}
\end{figure}

Having identified and validated the compact set of components that drive overconfidence, we now ask: can intervening on these components at inference time recalibrate the model's verbalized confidence?
We study two simple intervention strategies applied to the top-10 CMC components and evaluate their effect on the \emph{full dataset} using Expected Calibration Error (ECE)~\citep{pavlovic2025understanding}, Brier score~\citep{glenn1950verification}, and reliability curves\footnote{See illustrations in Appendix~\ref{app:calibration_metrics} and Appendix \ref{app:reliability}}. Importantly, although the intervention references are estimated from a small set of Bucket~1 examples, evaluation is performed on the full dataset, allowing us to test whether circuit-level intervention generalizes beyond the overconfident discovery subset.
Both strategies share the same deployment principle: a global reference is computed once from a small set of Bucket~1 examples and then applied uniformly at inference time. As a result, neither method requires constructing per-sample counterfactuals during deployment.
Calibration results before and after intervention are reported in Tables~\ref{tab:interv_popqa}, \ref{tab:interv_mmlu}, and \ref{tab:interv_nqopen}.
Reliability curves under different intervention strategies are presented in Figures~\ref{fig:reliability_curve} and \ref{fig:reliability_curve_qwen}.

\underline{Mean ablation} suppresses the contribution of a target component by replacing its activation at \textsc{pos\_end} with a global reference mean computed from corrupted-prompt activations:
\begin{equation}
  \hat{h}_{\text{out}}(x, \textsc{pos\_end})
    \;\leftarrow\;
    \mu_{\text{ref}}
    = \frac{1}{N}\sum_{i=1}^{N}
      h_{\text{out}}(x_{\text{corrupt}}^{(i)},\, \textsc{pos\_end}).
\end{equation}

\underline{Activation steering} provides a softer and tunable alternative. Instead of overwriting the activation, it shifts the target component along a precomputed overconfidence direction:
\begin{equation}
  \mathbf{v}_{\text{conf}}
    = \frac{1}{N}\sum_{i=1}^{N}
      \bigl[h_{\text{out}}(x_{\text{clean}}^{(i)},\, \textsc{pos\_end})
            - h_{\text{out}}(x_{\text{corrupt}}^{(i)},\, \textsc{pos\_end})\bigr].
\end{equation}
During confidence inference, we subtract a scaled version of it from each target activation:
\begin{equation}
  \hat{h}_{\text{out}}(x, \textsc{pos\_end})
    \;\leftarrow\;
    h_{\text{out}}(x, \textsc{pos\_end}) - \alpha \cdot \mathbf{v}_{\text{conf}},
\end{equation}
where $\alpha \in [0,1]$ controls the intervention strength. Unlike mean ablation, which fully replaces the activation, steering preserves the instance-specific signal while nudging it away from the overconfident direction, providing a continuous recalibration knob.

\textbf{Targeted intervention substantially improves calibration.}
Direct intervention on the top-ranked CMC components yields substantial calibration gains across models and datasets.
This improvement is evident both in aggregate metrics and in bin-level calibration behavior.
In terms of ECE and Brier score (Tables~\ref{tab:interv_popqa}, \ref{tab:interv_mmlu}, \ref{tab:interv_nqopen}), both mean ablation and activation steering are most effective on PopQA and NQOpen, where the best settings reduce ECE by roughly 78--97\% on PopQA and 81--83\% on NQOpen; improvements on MMLU are smaller but still consistent, reaching about 33\% for Qwen and 57\% for Llama.
%From an aggregate-metric perspective, both mean ablation and activation steering sharply reduce ECE and Brier score, with the strongest effects on PopQA and NQOpen: the best settings reduce ECE by roughly 78--97\% on PopQA and 81--83\% on NQOpen, while improvements on MMLU are smaller but still consistent, reaching about 33\% for Qwen and 57\% for Llama. 
Reliability curves provide a complementary and more intuitive view. For Llama-3.2-3B-Instruct (Figure~\ref{fig:reliability_curve}), both interventions move the confidence--accuracy curves substantially closer to the perfect calibration diagonal on PopQA and NQOpen. On MMLU, the curves improve more gradually, consistent with the weaker metric gains. Together, these results show that the discovered CMCs are not merely descriptive artifacts, but practical intervention targets for recalibration.

\textbf{Steering provides more precise and controllable recalibration than mean ablation.}
Both interventions improve calibration, but steering is more controllable.
Mean ablation fully overwrites the target activation, which can be highly effective when the overconfidence signal is tightly concentrated, but brittle otherwise: for example, it yields a 96.9\% ECE reduction for Llama/PopQA but only 3.1\% for Llama/MMLU.
Steering instead subtracts only the overconfidence direction while preserving the instance-specific residual.
As a result, it exhibits a consistent dose--response pattern: ECE improves from $\alpha = 0.3$ to an optimum around $0.5$--$0.6$, then degrades as stronger intervention suppresses useful confidence-related signals.
This optimal range is stable across four of six configurations, making $\alpha$ a practical hyperparameter requiring minimal tuning.
Steering therefore provides a more stable recalibration mechanism overall.
The full calibration curves over steering strengths are provided in Appendix~\ref{app:alpha_sweep}.

\textbf{Comparison with post-hoc calibration baselines.}
Appendix~\ref{app:baselines} compares our interventions with temperature scaling \citep{guo2017calibration}, Platt scaling \citep{platt1999probabilistic}, isotonic regression \citep{zadrozny2002transforming}, and histogram binning \citep{zadrozny2001obtaining}.
More specifically, our aim is not to outperform existing calibration methods but to establish circuit-level recalibration as a mechanistic complement that edits the internal computation rather than remapping outputs, and whose full-dataset ECE reduction from editing only the discovered components is itself further causal evidence for the discovered circuit.

\section{Discussion}
\textbf{Verbalized overconfidence likely reflects a mismatch between confidence verbalization and factual uncertainty.}
Our results can be interpreted in light of \citet{teplica2025sciurus}, which argues that a model’s actual uncertainty is generated largely in the same parts in LLMs that support factuality.
If verbal confidence faithfully reported this uncertainty, one would expect the circuit underlying verbal confidence to align closely with factual retrieval circuit and to vary with the task-dependent computation that supports answering. 
Instead, the circuit driving inflated verbal confidence is highly conserved across datasets within each model, and the answer-substitution controls (Application~\ref{app:robustness}) show that it responds to the committed answer largely independently of that answer's correctness, consistent with behavioral evidence that verbal confidence is largely answer-independent \citep{seo-etal-2026-advice}. 
The direct circuit comparison (Figure~\ref{fig:dual}) reinforces this mismatch: the verbal confidence circuit is more concentrated in middle-to-late layers and denser near the logits, whereas the factual retrieval circuit spans a longer pathway through earlier and middle layers. 
%This structural divergence suggests that verbalized confidence is not a faithful readout of the model’s query-specific uncertainty, but is instead shaped by a downstream confidence-writing mechanism that may impose a default high-confidence bias.
% Taken together, these findings suggest that verbalized overconfidence arises from a compact causal core that is structurally misaligned with the model's factuality-sensitive computation. The model encodes genuine uncertainty through its reasoning pathway, but the confidence verbalization circuit fails to faithfully relay that signal, and is instead shaped by a confidence-writing mechanism that may impose a default high-confidence bias
Taken together, these findings suggest that verbalized overconfidence arises from a compact, correctness-insensitive confidence-writing core that is structurally misaligned with the model's factuality-sensitive computation and may impose a default high-confidence bias.

\textbf{CMCs form a compact causal core of verbalized overconfidence, but not the full confidence-estimation system.}
The intervention results show that the discovered CMCs account for a substantial share of verbalized overconfidence: manipulating only the top-10 components already yields large calibration gains. 
At the same time, the best results occur at moderate steering strengths rather than under maximal intervention. 
One interpretation is that these components form only part of a broader confidence-estimation system, so stronger intervention begins to disrupt other useful confidence-related computation. 
Another is that the steering direction $\mathbf{v}_{\text{conf}}$, estimated as an average over Bucket~1 samples, only approximates the true per-example overconfidence axis, so excessive scaling amplifies this approximation error. 
Under either interpretation, the CMCs are best understood as a compact causal core that disproportionately biases the model toward inflated confidence, while the full system of confidence estimation remains an open question.

\textbf{Post-training may strengthen a prior toward overconfident behavior.}
The models studied here are instruction-tuned variants, raising the possibility that post-training contributes to the overconfidence mechanism we identify. Prior work shows that RLHF can exacerbate verbalized overconfidence and that aligned language models are often more overconfident than their pretrained counterparts~\citep{leng2025taming,he2023investigating}. A natural next step is to compare pretrained, supervised fine-tuned, and RLHF-tuned checkpoints, or to track how verbal confidence-related circuits emerge over the course of post-training.

%\paragraph{Limitations.}
%Our experiments are limited to two 3B-scale instruction-tuned models across three single-hop QA-style benchmarks, so it remains unclear how the identified circuits scale to larger models or more diverse settings such as multi-hop reasoning, long-form generation, and open-ended conversations.

\section{Related Work}
\textbf{LLM calibration and verbal confidence.}
LLMs can be prompted to verbalize confidence, sometimes with reasonable calibration~\citep{lin2022teaching,tian2023just}, but subsequent work found that verbalized confidence is often miscalibrated~\citep{xiong2024can,yona2024can}, a problem that post-training and RLHF can further exacerbate~\citep{he2023investigating,leng2025taming}.
We complement this literature by asking \emph{what internal mechanism causally drives} inflated verbal confidence, and whether intervening on it can improve calibration.

\textbf{Mechanistic interpretability}
seeks to reverse-engineer the internal circuits that support model behavior. Prior work has localized circuits for behaviors such as indirect object identification, mathematical reasoning, and factual/entity knowledge~\citep{wang2023interpretability,hanna2023does,ferrando2025do,conmy2023towards}.
More recent work has emphasized scalable or practical workflows for applications to larger models and more realistic tasks~\citep{rai2024practical,hanna2024have}.
We build on this line of work methodologically: our pipeline integrates EAP-IG~\citep{hanna2024have} for circuit attribution.

\textbf{Mechanistic studies of uncertainty and verbal confidence.}
\citet{teplica2025sciurus} studies the mechanistic basis of actual uncertainty and argues that it is generated in largely the same parts of the network that support factuality.
Relatedly, \citet{orgad2025llms} show that LLMs internally encode more truthfulness information than they externally reveal. 
On verbal confidence specifically, \citet{ji2025calibrating} characterize verbal uncertainty as a linear feature and intervene on it to reduce hallucinations. 
\citet{kumaran2026how} analyze how verbal confidence is computed and argue for a cached-retrieval account, while \citet{stolfo2024confidence} identify neurons that regulate next-token confidence more generally.
Our work focuses on a failure mode: the tendency of LLMs to be \emph{confidently wrong}.
We identify the compact circuit that drives this inflated verbalized confidence, validate its causal role, and show that targeted intervention on these circuits yields substantial calibration improvement.

\section{Conclusion}
We present a circuit-level mechanistic analysis of inflated verbalized confidence in LLMs.
By capturing output-level verbal confidence as a differentiable internal proxy, we identify compact \emph{Confidence Mover Circuits} that causally drive the model's overconfident behavior.
These circuits are causally validated and substantially conserved across datasets, suggesting that verbalized overconfidence is supported by a stable model-internal mechanism rather than a benchmark-specific artifact. We further show that targeted inference-time intervention on only a small set of top-ranked components substantially improves calibration, establishing that the discovered circuits are not only interpretable but also actionable. 
Overall, our results suggest that verbalized overconfidence is rooted in identifiable internal circuit, and that circuit-level intervention offers a promising complement to post-hoc calibration methods.

%\section*{Ethics Statement}
%Authors can add an optional ethics statement to the paper. 
%For papers that touch on ethical issues, this section will be evaluated as part of the review process. The ethics statement should come at the end of the paper. It does not count toward the page limit, but should not be more than 1 page.

\bibliographystyle{colm2026_conference}
\bibliography{colm2026_conference}

\appendix

%\section{Preliminaries and Setup}

\section{Models and Datasets}
\label{sec:model_data}
\textbf{Models.}
We experiment on two open-weight instruction-tuned model families with
distinct architectures:
\textbf{Qwen2.5-3B-Instruct}~\citep{qwen2.5} (36 layers, 16 query heads,
GQA with 2 KV heads) and
\textbf{Llama-3.2-3B-Instruct}~\citep{grattafiori2024llama} (28 layers,
24 query heads, GQA with 8 KV heads).

\textbf{Datasets.}
We evaluate on three knowledge-intensive QA benchmarks spanning different
question formats:
\textbf{PopQA}~\citep{mallen-etal-2023-trust} (14{,}267 open-domain
factoid questions with associated Wikipedia-popularity scores),
\textbf{MMLU}~\citep{hendrycks2021measuring} (14{,}042 multiple-choice knowledge
questions across 57 subjects), and
\textbf{NQOpen}~\citep{lee-etal-2019-latent} (3{,}610 open-domain questions
derived from Google search queries).
PopQA and NQOpen test free-form factual recall; MMLU introduces a
constrained multiple-choice format.
For each model$\times$dataset combination, we run the two-step elicitation pipeline (\S\ref{sec:tsld}), retain only incorrect predictions, and stratify into three buckets for circuit discovery.

\section{LLM as a Computational Graph}
\label{sec:llm_cg}
We conceptualize the LLM as a Directed Acyclic Graph (DAG), denoted as $\mathcal{G} = (\mathcal{V}, \mathcal{E})$, following the standard view in circuit-level mechanistic interpretability. In this graph, nodes $\mathcal{V}$ correspond to computational modules, and edges 
$\mathcal{E}$ represent information flow between modules through the residual stream. This formulation provides a convenient abstraction for identifying the subgraph that causally supports a target behavior.

\textbf{Nodes ($\mathcal{V}$).} 
To enable fine-grained circuit discovery, we decompose the standard Transformer blocks into the atomic linear projections of the models:
\begin{equation}
\mathcal{V} = \{ W_Q^{(l,h)}, W_K^{(l,h)}, W_V^{(l,h)}, W_O^{(l,h)}, W_{MLP}^{(l)} \mid l \in [1, L], h \in [1, H] \}
\end{equation}
where $L$ is the number of layers and $H$ is the number of heads per layer. $W_Q$, $W_K$, $W_V$ represent the query, key, and value projections for the $h$-th head in layer $l$, $W_O$ is the output projection, and $W_{MLP}$ represents the two-layer feed-forward network.

\textbf{Edges ($\mathcal{E}$).}
An edge $e_{u \to v}$ exists if the output of module $u$ contributes to the input of module $v$ via the residual stream.
Intuitively, edges capture how intermediate representations are transmitted and transformed across the network. Our goal is to identify the subset of edges whose activations are causally responsible for the target behavior of interest, namely inflated verbalized confidence.

\textbf{Behavioral metric.}
Given an input $x$, let $f(x)$ denote the scalar behavior of interest. In this work, $f(x)$ is instantiated as the Target-Set Logit Difference (TSLD) defined in Section~\ref{sec:tsld}, which serves as a differentiable proxy for verbalized confidence. Circuit attribution is therefore performed with respect to changes in this scalar signal.
\section{Edge Attribution Patching with Integrated Gradients (EAP-IG)}
\label{sec:eapig_prelim}
To identify the subgraph responsible for a target behavior, we require a method that attributes changes in the behavioral metric to specific internal edges. Standard methods like Activation Patching~\citep{meng2022locating} are faithful but computationally expensive, while vanilla gradient attribution is efficient but can be unreliable in deep nonlinear networks due to gradient saturation. To balance fidelity and efficiency, we employ Edge Attribution Patching with Integrated Gradients (EAP-IG)~\citep{hanna2024have}.

Vanilla EAP~\citep{neelattribution} approximates the importance of an edge $e$ using the product of its activation $x_e$ and the gradient of the scalar objective $\mathcal{L}$ with respect to that activation:
\begin{equation}
    \phi_{EAP} = x_e\cdot \frac{\partial \mathcal{L}}{\partial x_e}
\end{equation}
where, in our setting, $\mathcal{L}$ is the confidence-related signal induced by TSLD.
While requiring only one backward pass, this method can fail under saturation effect. When a neuron is fully activated, the local gradient is near zero, even if that neuron is critically important for the output, leading to incomplete or fragmented circuits.

%\subsection{EAP with Integrated Gradients}
To overcome saturation, EAP-IG~\citep{hanna2024have} accumulates gradients along a linear path between a corrupted baseline input ($I_{corrupt}$) and the clean target input ($I_{clean}$).
Let $x_e$ denote the activation along edge $e$. A path is defined as $x_e(\gamma) = x_e^{corrupt} + \gamma(x_e^{clean} - x_e^{corrupt})$ for $\gamma \in [0,1]$.
The attribution score $\phi(e)$ for an edge $e$ is calculated as:
\begin{equation}
    \phi(e) = (x_e^{clean} - x_e^{corrupt}) \times \int_{\gamma=0}^{1} \frac{\partial \mathcal{L}(x(\gamma))}{\partial x_e(\gamma)} d\gamma
\end{equation}
By accumulating gradients along the interpolation path rather than relying on a single local derivative, EAP-IG yields more reliable edge attributions under saturation.
In practice, this integral is approximated using a Riemann sum with $m$ steps:
\begin{equation}\label{eq:eapig_approx}
    \phi(e) \approx (x_e^{clean} - x_e^{corrupt}) \times \frac{1}{m} \sum_{k=1}^{m} \frac{\partial \mathcal{L}(x(\frac{k}{m}))}{\partial x_e}
\end{equation}
%This formalism provides a robust, continuous estimate of causal importance, enabling us to map the full reasoning circuitry even through saturated MLP layers.

\section{Raw Verbal Confidence Distributions}

\label{app:conf_dist}
\begin{figure}[h]
    \centering
    \includegraphics[width=\textwidth]{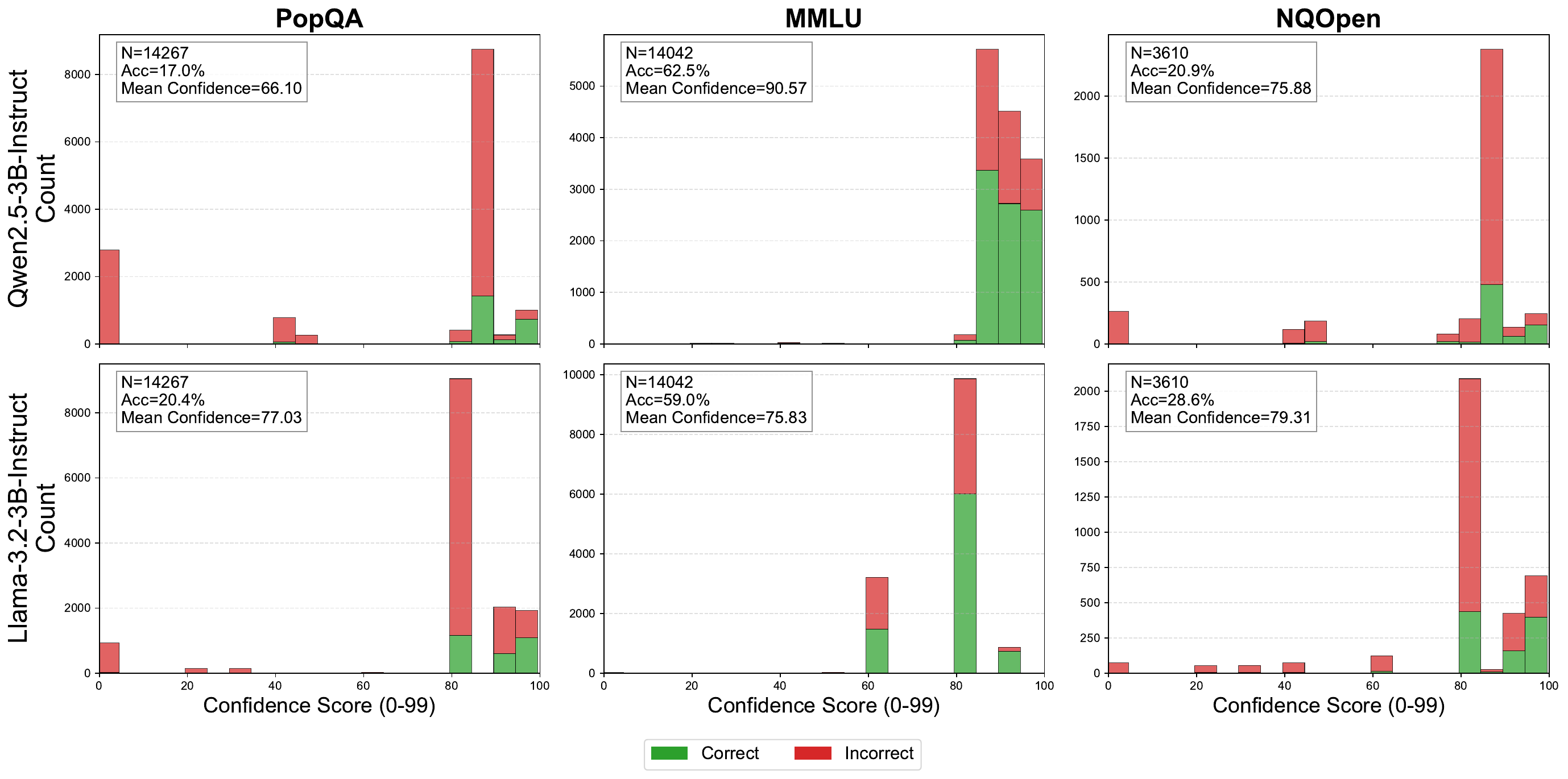}
    \caption{
    \textbf{Raw verbalized confidence distributions} across all six
    model$\times$dataset configurations.
    Green: correct predictions; red: incorrect predictions.
    }
    \label{fig:conf}
\end{figure}

Figure~\ref{fig:conf} shows the distribution of verbalized confidence scores
across all six model$\times$dataset configurations, separated by answer correctness.
These distributions vividly illustrate the core phenomenon studied in this paper: the models are not merely wrong, but \emph{confidently wrong}, systematically assigning high confidence scores that poorly discriminate between correct and incorrect answers.

\section{Robustness Controls for Circuit Discovery}
\label{app:robustness}

\subsection{Answer-Substitution Control}
\label{app:answer_substitution}

The truth-injection counterfactual replaces the model's
incorrect Step~1 answer with the ground-truth answer.
Although this intervention changes the factual status of
the committed answer, it may also alter semantic content,
surface form, token identity, and sequence length.

To test whether these factors account for the recovered
circuit, we repeat EAP-IG circuit discovery on
Qwen2.5-3B-Instruct/PopQA using four Step~1 answer
substitutions:

\begin{enumerate}
    \item the ground-truth answer used in the main
    experiment;

    \item a same-meaning paraphrase, implemented as a
    different valid PopQA gold alias;

    \item a type-matched plausible-wrong answer,
    sampled from another answer associated with the
    same relation type; and

    \item a length-matched distractor with comparable
    token length but unrelated semantic content.
\end{enumerate}

For example, when the question is
``What is Henry Feilden's occupation?''
and the model answers ``architect,'' the four
substitutions are ``politician,'' ``political leader,''
``priest,'' and ``board game,'' respectively.

We evaluate each condition using the mean
$\Delta_{\text{TSLD}}$ and the overlap between its
top-10 attributed components and those recovered under
the original ground-truth substitution.

\begin{table}[h]
\centering
\small

\begin{tabular}{lcc}
\toprule
Substituted Step~1 answer
& Mean $\Delta_{\text{TSLD}}$
& Top-10 overlap \\
\midrule
Ground truth
& $-8.50$
& 10/10 \\

Same-meaning gold alias
& $-8.24$
& 9/10 \\

Type-matched plausible-wrong answer
& $-7.71$
& 10/10 \\

Length-matched distractor
& $-8.68$
& 7/10 \\
\bottomrule
\end{tabular}

\caption{
  \textbf{Answer-substitution controls on
  Qwen2.5-3B-Instruct/PopQA.}
  Mean $\Delta_{\text{TSLD}}$ measures the change
  relative to the original incorrect-answer context,
  with more negative values indicating a larger
  confidence decrease.
  Top-10 overlap is measured against the circuit
  recovered using ground-truth substitution.
}
\label{tab:answer_substitution_controls}
\end{table}

% The same-meaning alias recovers nine of the ten
% components identified under ground-truth substitution:
% M30, M35, M26, M31, M25, M27, M28, M24, and L24H5.

% The type-matched plausible-wrong condition recovers all
% ten components:
% M30, M35, M26, M31, M25, M27, M28, M0, M24, and
% L24H5.

% The length-matched distractor recovers seven components:
% M30, M25, M35, M31, L24H5, M26, and M28.

As shown in Table~\ref{tab:answer_substitution_controls}, all four substitutions produce confidence decreases of
similar magnitude and recover substantially overlapping
component sets. The same-meaning alias result shows that
the circuit is not tied to the exact surface form of the
ground-truth answer. The length-matched control indicates
that answer length alone does not explain the attribution
pattern.
More importantly, the plausible-wrong condition produces
a confidence shift and component ranking comparable to
those obtained with the ground truth. Therefore, the
discovered circuit cannot be interpreted as selectively
detecting whether the substituted answer is factually
correct.

The results instead support a more precise interpretation:
the CMC writes verbalized confidence conditioned on the
model's committed answer, while remaining largely
insensitive to whether that answer is factually correct.
That is, the model is confident in what it said, not in what is true.

% These controls are limited to
% Qwen2.5-3B-Instruct/PopQA. Their replication across
% additional models, datasets, and answer formats remains
% future work.

\subsection{Elicitation-Prompt Control}
\label{app:prompt_robustness}

The main experiments use a fixed Step~2 prompt to elicit
verbalized confidence. To test whether the recovered
components depend on this exact wording, we repeat
circuit discovery on Qwen2.5-3B-Instruct/PopQA using
three semantically related prompt variants.

\begin{table}[h]
\centering
\small

\resizebox{\linewidth}{!}{
\begin{tabular}{
  l
  c
  c
  c
}
\toprule
Step~2 confidence prompt
&
Mean $\Delta_{\text{TSLD}}$
&
Top-5 overlap
&
Top-10 overlap
\\
\midrule

Original prompt
&
$-7.07$
&
5/5
&
10/10
\\

``How certain are you that it is correct?''
&
$-10.63$
&
4/5
&
8/10
\\

``On a scale from 0 to 100, how sure are you?''
&
$-3.75$
&
3/5
&
7/10
\\

``How likely is it that your answer is accurate?''
&
$-3.35$
&
4/5
&
6/10
\\

\bottomrule
\end{tabular}
}

\caption{
  \textbf{Confidence-prompt robustness on
  Qwen2.5-3B-Instruct/PopQA.}
  The circuit remains partially conserved under three
  semantically related Step~2 prompt variants.
}
\label{tab:prompt_robustness}
\end{table}

The top-ranked components remain conserved
across all prompt variants, suggesting that the discovered circuits are properties of the models rather than artifacts of a single elicitation template.

\subsection{TSLD Set Sensitivity}
\label{app:tsld_sensitivity}

We vary the high- and low-confidence sets used in Eq.~\ref{eq:tsld} and measure the Jaccard similarity between the resulting top-10 components and those discovered with the default sets (Table~\ref{tab:tsld_sensitivity}).

\begin{table}[h]
\centering

\resizebox{\linewidth}{!}{
\begin{tabular}{
  l
  c
}
\toprule
High/low target sets
&
Top-10 Jaccard vs.\ default
\\
\midrule

$\mathcal{H}=\{70,75,80,85,90,99\}$;
$\mathcal{L}=\{0,10,15,20,25,30\}$
&
1.00
\\

$\mathcal{H}=\{80,85,90,99\}$;
$\mathcal{L}=\{0,10,20\}$
&
1.00
\\

$\mathcal{H}=
\{60,65,70,75,80,85,90,95,99\}$;
$\mathcal{L}=
\{0,5,10,15,20,25,30,35,40\}$
&
0.82
\\

$\mathcal{H}=\{80,90,99\}$;
$\mathcal{L}=\{0,15,30\}$
&
0.82
\\

$\mathcal{H}=\{75,80,85,90,95,99\}$;
$\mathcal{L}=\{5,10,15,20,25,35\}$
&
0.82
\\

\bottomrule
\end{tabular}
}

\caption{
  \textbf{Sensitivity of
  Qwen2.5-3B-Instruct/PopQA circuit discovery
  to TSLD target sets.}
  The recovered top-10 components remain stable
  across narrower, wider, coarser, and shifted
  target sets.
}
\label{tab:tsld_sensitivity}
\end{table}

\subsection{Bucket-2 Contrast}
Running the same EAP-IG discovery on Bucket~2 (the opposite-sign contrast in which truth injection increases confidence) yields an attention-heavy circuit whose top components (L26H3, L28H4, M13, M19, L35H3, and others) share none of the Bucket~1 top-10 (Jaccard 0.00).
A distinct circuit is expected for an opposite-sign phenomenon; the disjointness confirms that the Bucket~1 circuit is specific to overconfident collapse rather than a by-product of the stratification.

\section{Calibration Metrics}
\label{app:calibration_metrics}

We evaluate recalibration using two standard metrics.

\textbf{Expected Calibration Error (ECE)}~\citep{pavlovic2025understanding}
measures the average gap between predicted confidence and observed accuracy.
Predictions are partitioned into $B$ equal-width bins by confidence score,
and ECE is computed as the weighted average of per-bin calibration gaps:
\begin{equation}
  \text{ECE} = \sum_{b=1}^{B} \frac{n_b}{N}
    \bigl|\text{acc}(b) - \text{conf}(b)\bigr|,
\end{equation}
where $n_b$ is the number of samples in bin $b$, $N$ is the total number of
samples, $\text{acc}(b)$ is the empirical accuracy within the bin, and
$\text{conf}(b)$ is the mean predicted confidence.
A perfectly calibrated model has ECE $= 0$.
We use $B = 10$ bins throughout.

\textbf{Brier Score}~\citep{glenn1950verification} provides a complementary
assessment by measuring the mean squared error between the predicted
confidence (normalized to $[0,1]$) and the binary correctness indicator:
\begin{equation}
  \text{Brier} = \frac{1}{N} \sum_{i=1}^{N}
    \bigl(c_i - y_i\bigr)^2,
\end{equation}
where $c_i \in [0,1]$ is the predicted confidence for sample $i$ and
$y_i \in \{0, 1\}$ indicates whether the answer is correct.
Unlike ECE, which captures only calibration, the Brier score jointly
penalizes miscalibration and poor discrimination (inability to separate
correct from incorrect predictions), providing a more holistic measure of
confidence quality.
Lower values indicate better performance for both metrics.

\section{Reliability Curves}
\label{app:reliability}

Reliability curves provide a bin-level view of calibration. They plot \emph{empirical accuracy} against \emph{mean predicted confidence} over confidence bins, with the diagonal corresponding to perfect calibration: if the model reports confidence \(p\), it should be correct approximately \(p\) fraction of the time. Curves below the diagonal indicate \emph{overconfidence}, meaning that the model is less accurate than its stated confidence suggests, while curves above the diagonal indicate \emph{underconfidence}.
Compared with aggregate metrics such as ECE and Brier score, reliability curves provide a more fine-grained and interpretable view of calibration by showing where miscalibration occurs across the confidence range and whether it takes the form of overconfidence or underconfidence.
The shaded area between the curve and the diagonal visualizes the total calibration gap, which ECE summarizes as a single scalar. Intuitively, the closer the curve lies to the diagonal, the better calibrated the model is.

\begin{figure}[h]
    \centering
    \includegraphics[width=\textwidth]{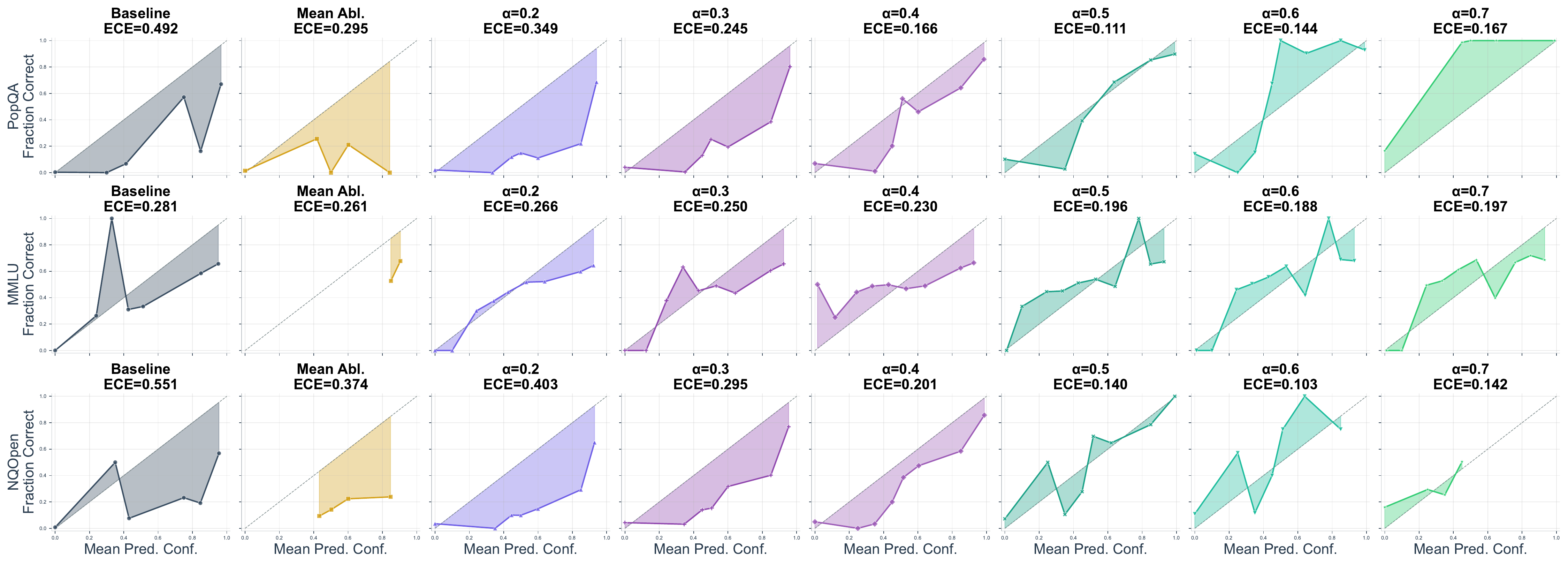}
    \caption{
    \textbf{Reliability curves for Qwen2.5-3B-Instruct} across PopQA (top), MMLU (middle), and NQOpen (bottom) under baseline (before intervention), mean ablation, and steering at $\alpha \in \{0.2, 0.3, 0.4, 0.5, 0.6, 0.7\}$.
    }
    \label{fig:reliability_curve_qwen}
\end{figure}

Figures~\ref{fig:reliability_curve_qwen} and~\ref{fig:reliability_curve} show
reliability curves for Qwen2.5-3B-Instruct and Llama-3.2-3B-Instruct respectively, across all three datasets under baseline (before intervention), mean ablation, and
steering at $\alpha \in \{0.2, \ldots, 0.7\}$.

\section{Full Steering Strength Sweep from $0.1$ to $1$}
\label{app:alpha_sweep}

\begin{figure}[h]
    \centering
    \includegraphics[width=\textwidth]{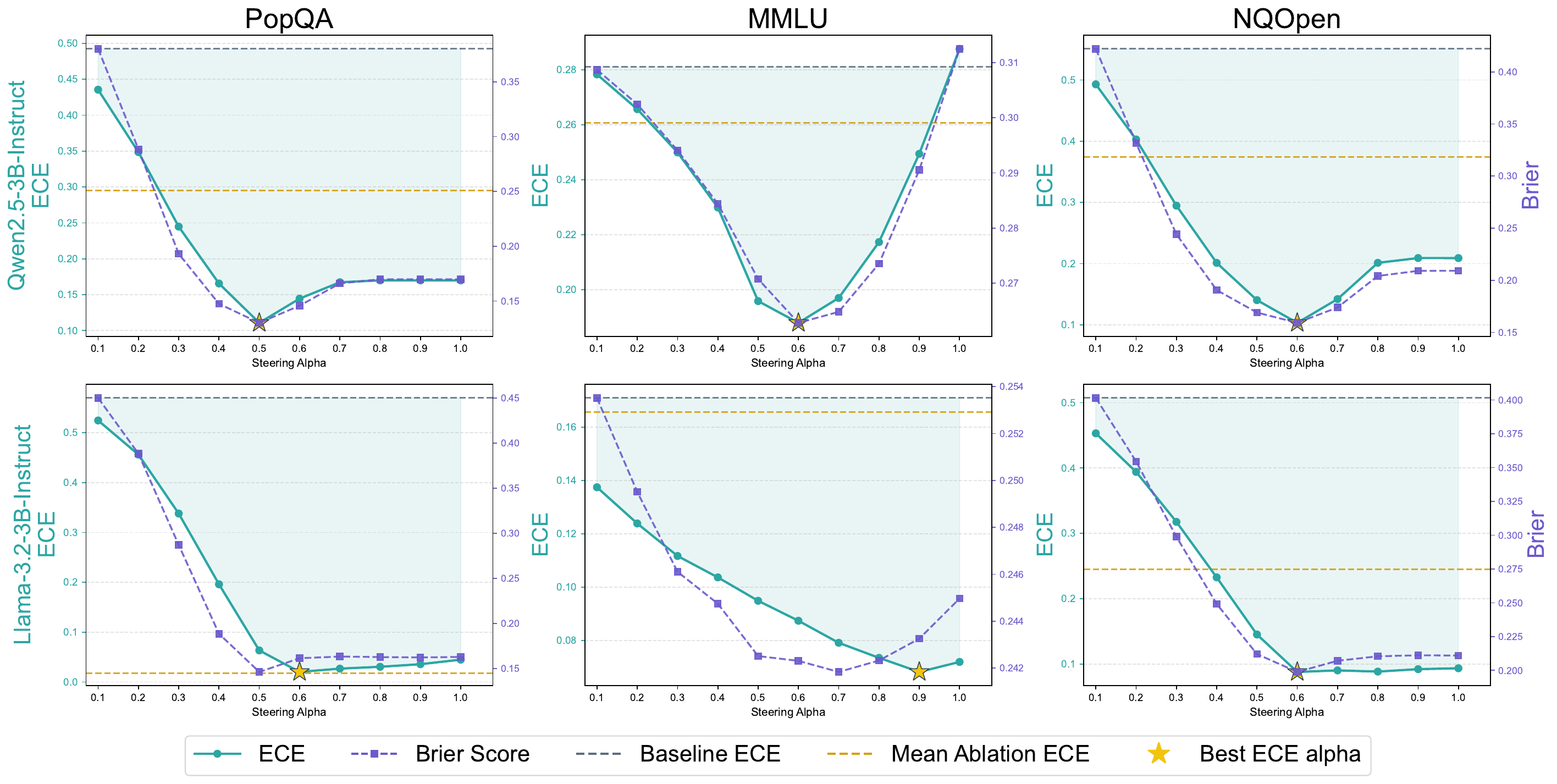}
    \caption{\textbf{Steering $\alpha$ sweep across all six configurations.}
    ECE (teal, left axis) and Brier score (purple, right axis) as a function
    of steering strength $\alpha$.
    Dashed grey: baseline ECE; dashed yellow: mean ablation ECE; star:
    optimal $\alpha$.
    }
    \label{fig:steer}
\end{figure}

Figure~\ref{fig:steer} shows ECE (left axis) and Brier score (right axis)
as a function of steering strength $\alpha$.
Across all six model$\times$dataset settings, steering exhibits a clear dose-response pattern: both ECE and Brier score improve substantially as $\alpha$ increases from small values, indicating that progressively stronger suppression of the overconfidence direction yields better calibration. However, this improvement is not monotonic. In most settings, performance reaches an optimum at a moderate intervention strength and then plateaus or degrades as $\alpha$ becomes too large.
Overall, the sweep confirms that activation steering provides a robust and tunable recalibration mechanism. It also shows that the intervention is not simply “the stronger, the better”: calibration improves most when the overconfidence direction is attenuated to an intermediate degree rather than fully suppressed.

\section{Comparison with Post-Hoc Calibration Baselines}
\label{app:baselines}
We compare circuit-level intervention against four standard output-level calibrators: temperature scaling \citep{guo2017calibration}, Platt scaling \citep{platt1999probabilistic}, isotonic regression \citep{zadrozny2002transforming}, and histogram binning \citep{zadrozny2001obtaining}.
Each calibrator is fit on a labeled 50\% split of the verbalized confidence scores and evaluated on the disjoint 50\% split; the Raw ECE column is computed on that evaluation split and therefore deviates slightly from the full-dataset baselines in Tables~\ref{tab:interv_popqa}, \ref{tab:interv_mmlu}, and \ref{tab:interv_nqopen}.
The mean-ablation and best-steering columns restate our full-dataset results from those tables.
The comparison is included to position circuit-level recalibration as a viable, mechanistic complement to output-level post-hoc calibration methods, rather than to claim state-of-the-art calibration performance

\begin{table}[h]
\centering
\small
\setlength{\tabcolsep}{4pt}

\resizebox{\linewidth}{!}{
\begin{tabular}{lccccccc}
\toprule
Model / Dataset
& Raw ECE
& Temp.
& Platt
& Isotonic
& Histogram
& Mean abl.\ (ours)
& Best steer (ours)
\\
\midrule

Llama-3.2-3B / PopQA
& 0.568
& 0.302
& 0.008
& 0.075
& 0.147
& 0.018
& 0.020
\\

Qwen2.5-3B / NQOpen
& 0.555
& 0.304
& 0.016
& 0.054
& 0.033
& 0.374
& 0.103
\\

Llama-3.2-3B / MMLU
& 0.176
& 0.021
& 0.026
& 0.078
& 0.232
& 0.166
& 0.074
\\

Llama-3.2-3B / NQOpen
& 0.507
& 0.245
& 0.036
& 0.057
& 0.191
& 0.245
& 0.088
\\

\bottomrule
\end{tabular}
}

\caption{
  \textbf{Comparison with post-hoc calibrators}
  (ECE, 10 bins; lower is better).
  Calibrators are fit on a labeled 50\% split and
  evaluated on the disjoint half; Raw ECE is computed
  on the evaluation split.
  Mean ablation and best steering restate our
  full-dataset results from
  Tables~\ref{tab:interv_popqa},
  \ref{tab:interv_mmlu}, and
  \ref{tab:interv_nqopen}.
}
\label{tab:posthoc_baselines}
\end{table}

\section{Model Scale Generalization}
\label{app:scale}
\begin{table}[h]
\centering
\small
\begin{tabular}{ccc}
  \toprule
  Rank & PopQA & NQOpen \\
  \midrule
  1  & m27    & m27 \\
  2  & m21    & m21 \\
  3  & m23    & m26 \\
  4  & m26    & m23 \\
  5  & m20    & m20 \\
  6  & m19    & m19 \\
  7  & m25    & m25 \\
  8  & m18    & m24 \\
  9  & m24    & m22 \\
  10 & a20.h1 & m18 \\
  \bottomrule
\end{tabular}
\caption{\textbf{Top-10 components by attribution score for Qwen2.5-7B-Instruct.} 9 of the top-10 components are shared across PopQA and NQOpen.}
\label{tab:scale_discovery}
\end{table}
To test whether the discovered mechanism persists at larger scale within the same model family, we repeat circuit discovery and the steering sweep on \textbf{Qwen2.5-7B-Instruct} for PopQA and NQOpen.
Table~\ref{tab:scale_discovery} lists the top-10 components: 9 of the top-10 are shared across the two datasets, and the circuit again consists of mid-to-late MLP blocks, the same locus as at 3B.
Table~\ref{tab:scale_sweep} reports the steering sweep: at $\alpha=0.6$, ECE falls from 0.647 to 0.068 on PopQA and from 0.626 to 0.109 on NQOpen, with the same dose--response pattern as at 3B, while mean ablation is effective on PopQA (0.155) but not on NQOpen (0.615), mirroring the brittleness observed at 3B.
This cross-scale experimental result further complements the cross-family Qwen2.5-3B vs Llama-3.2-3B result.

\begin{table}[h]
\centering
\small
\begin{tabular}{lccccccc}
  \toprule
  Dataset & Baseline & $\alpha=0.2$ & $\alpha=0.3$ & $\alpha=0.4$ & $\alpha=0.5$ & $\alpha=0.6$ & Mean ablation \\
  \midrule
  PopQA  & 0.647 & 0.419 & 0.312 & 0.203 & 0.121 & \textbf{0.068} & 0.155 \\
  NQOpen & 0.626 & 0.528 & 0.450 & 0.347 & 0.217 & \textbf{0.109} & 0.615 \\
  \bottomrule
\end{tabular}
\caption{\textbf{Intervention results (ECE) for Qwen2.5-7B-Instruct.} Steering sweep over $\alpha$ and mean ablation; best ECE per dataset in \textbf{bold}.}
\label{tab:scale_sweep}
\end{table}

\begin{table}[h]
\centering
\small
\begin{tabular}{cll}
  \toprule
  Bucket & Criterion & Role \\
  \midrule
  1 & $\Delta_{\text{TSLD}} \leq -\tau$ & Overconfident collapse; circuit discovery target \\
  2 & $\Delta_{\text{TSLD}} \geq +\tau$ & Truth recognition; excluded \\
  3 & $\Delta_{\text{TSLD}} \approx 0$  & Noise; excluded \\
  \bottomrule
\end{tabular}
\caption{Three-bucket stratification.}
\label{tab:three_bucket}
\end{table}

\begin{table}[h]
\centering
\small
\definecolor{baselinegray}{HTML}{EFEFEF}
\begin{tabular}{ccccccc}
\toprule
Model & Method & $\alpha$ & ECE & Brier & ECE Impr. & Brier Impr. \\
\midrule
\multirow{8}{*}{\rotatebox[origin=c]{90}{{\scriptsize Qwen2.5-3B-Instruct}}}
  & \cellcolor{baselinegray}baseline      & \cellcolor{baselinegray}--  & \cellcolor{baselinegray}0.281          & \cellcolor{baselinegray}0.310 & \cellcolor{baselinegray}--            & \cellcolor{baselinegray}--          \\
  & mean ablation & --  & 0.261          & 0.298 & 7.3\%           & 3.8\%         \\
  & steering      & 0.3 & 0.250          & 0.294 & 11.1\%          & 5.2\%         \\
  & steering      & 0.4 & 0.230          & 0.284 & 18.2\%          & 8.3\%         \\
  & steering      & 0.5 & 0.196          & 0.271 & 30.3\%          & 12.7\%        \\
  & steering      & 0.6 & \textbf{0.188} & \textbf{0.263} & \textbf{33.1\%} & \textbf{15.3\%}        \\
  & steering      & 0.7 & 0.197          & 0.265 & 29.9\%          & 14.6\%        \\
  & steering      & 0.8 & 0.217          & 0.274 & 22.7\%          & 11.8\%        \\
\midrule
\multirow{8}{*}{\rotatebox[origin=c]{90}{{\scriptsize Llama-3.2-3B-Instruct}}}
  & \cellcolor{baselinegray}baseline      & \cellcolor{baselinegray}--  & \cellcolor{baselinegray}0.171          & \cellcolor{baselinegray}0.265 & \cellcolor{baselinegray}--            & \cellcolor{baselinegray}--          \\
  & mean ablation & --  & 0.166          & 0.269 & 3.1\%           & $-1.8$\%      \\
  & steering      & 0.3 & 0.112          & 0.246 & 34.7\%          & 7.0\%         \\
  & steering      & 0.4 & 0.104          & 0.245 & 39.4\%          & 7.5\%         \\
  & steering      & 0.5 & 0.095          & 0.243 & 44.5\%          & 8.3\%         \\
  & steering      & 0.6 & 0.087          & 0.242 & 48.9\%          & 8.4\%         \\
  & steering      & 0.7 & 0.079          & \textbf{0.242} & 53.7\%          & \textbf{8.6\%}         \\
  & steering      & 0.8 & \textbf{0.074} & 0.242 & \textbf{57.0\%} & 8.4\%         \\
\bottomrule
\end{tabular}%
\caption{\textbf{Intervention results on MMLU.}
  Best ECE and Brier per model in \textbf{bold}.}
\label{tab:interv_mmlu}
\end{table}

\begin{table}[h]
\centering
\small
\definecolor{baselinegray}{HTML}{EFEFEF}
\begin{tabular}{ccccccc}
\toprule
Model & Method & $\alpha$ & ECE & Brier & ECE Impr. & Brier Impr. \\
\midrule
\multirow{8}{*}{\rotatebox[origin=c]{90}{{\scriptsize Qwen2.5-3B-Instruct}}}
  & \cellcolor{baselinegray}baseline      & \cellcolor{baselinegray}--  & \cellcolor{baselinegray}0.551          & \cellcolor{baselinegray}0.485 & \cellcolor{baselinegray}--            & \cellcolor{baselinegray}--          \\
  & mean ablation & --  & 0.374          & 0.304 & 32.1\%          & 37.3\%        \\
  & steering      & 0.3 & 0.295          & 0.244 & 46.5\%          & 49.6\%        \\
  & steering      & 0.4 & 0.201          & 0.191 & 63.5\%          & 60.7\%        \\
  & steering      & 0.5 & 0.141          & 0.169 & 74.5\%          & 65.1\%        \\
  & steering      & 0.6 & \textbf{0.103} & \textbf{0.159} & \textbf{81.2\%} & \textbf{67.2\%}        \\
  & steering      & 0.7 & 0.142          & 0.174 & 74.2\%          & 64.1\%        \\
  & steering      & 0.8 & 0.201          & 0.204 & 63.4\%          & 57.9\%        \\
\midrule
\multirow{8}{*}{\rotatebox[origin=c]{90}{{\scriptsize Llama-3.2-3B-Instruct}}}
  & \cellcolor{baselinegray}baseline      & \cellcolor{baselinegray}--  & \cellcolor{baselinegray}0.507          & \cellcolor{baselinegray}0.453 & \cellcolor{baselinegray}--            & \cellcolor{baselinegray}--          \\
  & mean ablation & --  & 0.245          & 0.273 & 51.7\%          & 39.8\%        \\
  & steering      & 0.3 & 0.318          & 0.299 & 37.4\%          & 34.1\%        \\
  & steering      & 0.4 & 0.233          & 0.249 & 54.1\%          & 45.0\%        \\
  & steering      & 0.5 & 0.145          & 0.212 & 71.3\%          & 53.2\%        \\
  & steering      & 0.6 & \textbf{0.088} & \textbf{0.199} & \textbf{82.6\%} & \textbf{56.1\%}        \\
  & steering      & 0.7 & 0.091          & 0.207 & 82.2\%          & 54.3\%        \\
  & steering      & 0.8 & 0.089          & 0.210 & 82.5\%          & 53.6\%        \\
\bottomrule
\end{tabular}%
\caption{\textbf{Intervention results on NQOpen.}
  Best ECE and Brier per model in \textbf{bold}.}
\label{tab:interv_nqopen}
\end{table}

\begin{figure}[h]
    \centering
    \includegraphics[width=0.9\textwidth]{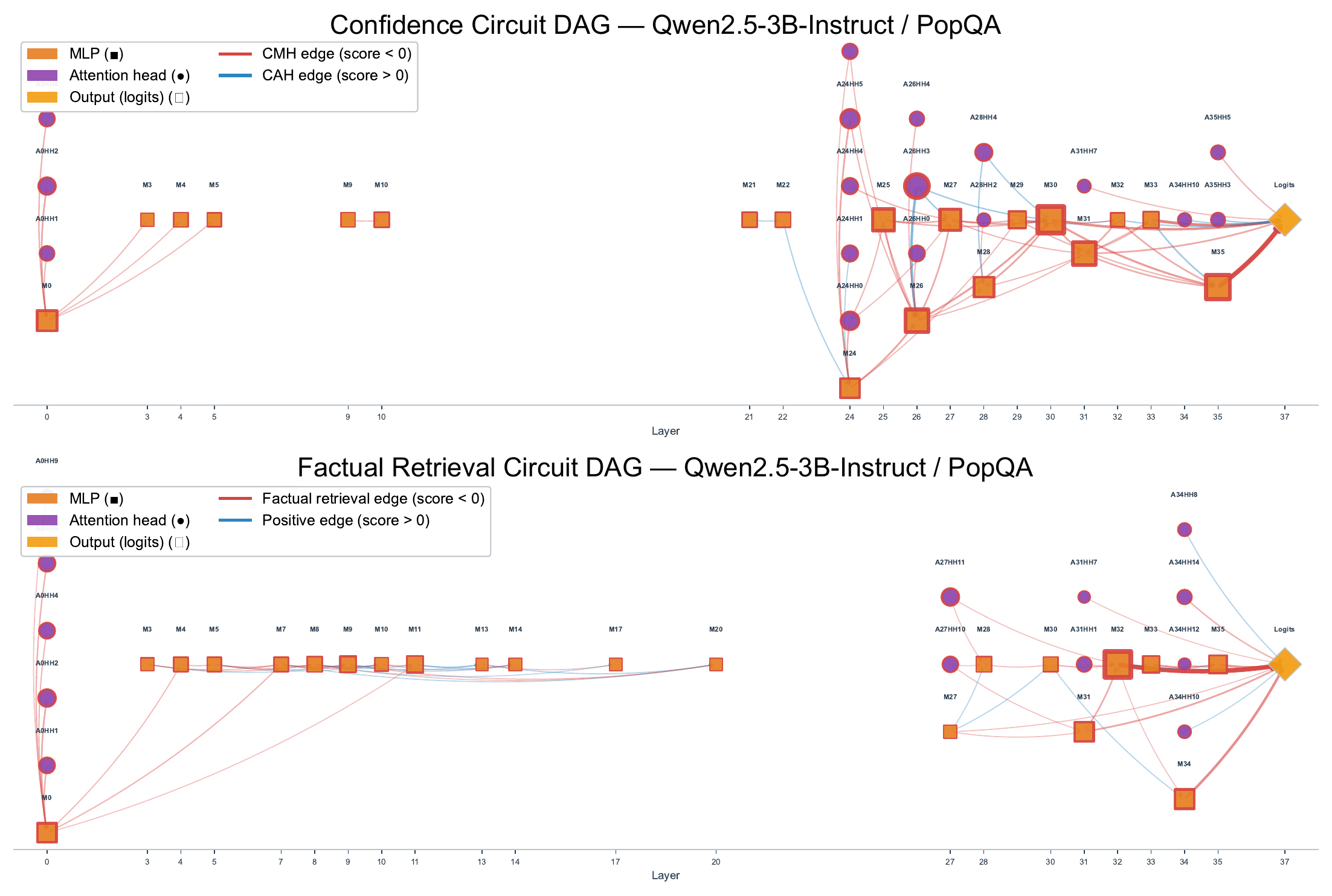}
    \caption{Verbal confidence circuit vs.\ factual retrieval circuit (Qwen2.5-3B-Instruct, PopQA).
    Both circuits are discovered on the same Bucket~1 records with the same EAP-IG machinery; the factual retrieval circuit uses an answer-logit objective in place of TSLD. The two circuits are structurally distinct: the Jaccard overlap of their top-10 (top-20) components is 0.18 (0.21), and the cosine similarity between their signed edge-attribution scores on the induced subgraphs is 0.004 (0.055).
    }
    \label{fig:dual}
\end{figure}

\end{document}